\documentclass[lettersize,journal]{IEEEtran}
\usepackage{amsmath,amssymb,amsfonts}
\usepackage{algorithmic}
\usepackage{array}
\usepackage{textcomp}
\usepackage{stfloats}
\usepackage{url}
\usepackage{verbatim}
\usepackage{graphicx}
\usepackage{cite}
\usepackage{cancel}
\usepackage{amsthm}
\usepackage[normalem]{ulem}
\usepackage[ruled,linesnumbered,noend]{algorithm2e}
\usepackage[colorlinks=true, allcolors=blue]{hyperref}
\usepackage{subcaption}
\DeclareMathOperator*{\argmax}{arg\,max}

\newcommand{\vb}[1]{\textbf{#1}}
\newcommand{\norm}[1]{\left\lVert#1\right\rVert}

\newcommand{\mt}[1]{\mathrm{#1}}

\newtheorem{theorem}{Theorem}[section]

\newtheorem{lemma}[theorem]{Lemma}

\theoremstyle{definition}

\newtheorem{assum}{Assumption}

\begin{document}

\title{Guaranteed Encapsulation of Targets with Unknown Motion by a Minimalist Robotic Swarm}

\author{Himani Sinhmar, Hadas Kress-Gazit
        % <-this % stops a space
\thanks{The authors are with the Sibley School of Mechanical and Aerospace Engineering, Cornell University, Ithaca, NY, 14853 USA. {\tt\small \{hs962,hadaskg\}@cornell.edu}. This work is supported by NSF EFMA-1935252.}% <-this % stops a space
}
\def\journalname{Transactions on Robotics}
\markboth{\journalname, VOL. XX, NO. XX, XXXX}%
{Author \MakeLowercase{\textit{Sinhmar and Kress-Gazit}}: Guaranteed Encapsulation of Targets with Unknown Motion by a Minimalist Robotic Swarm}


\maketitle

\begin{abstract}
We present a decentralized control algorithm for a robotic swarm given the task of encapsulating static and moving targets in a bounded unknown environment. We consider minimalist robots without memory, explicit communication, or localization information. The state-of-the-art approaches generally assume that the robots in the swarm are able to detect the relative position of neighboring robots and targets in order to provide convergence guarantees. In this work, we propose a novel control law for the guaranteed encapsulation of static and moving targets while avoiding all collisions, when the robots do not know the exact relative location of any robot or target in the environment. We make use of the Lyapunov stability theory to prove the convergence of our control algorithm and provide bounds on the ratio between the target and robot speeds. Furthermore, our proposed approach is able to provide stochastic guarantees under the bounds that we determine on task parameters for scenarios where a target moves faster than a robot. Finally, we present an analysis of how the emergent behavior changes with different parameters of the task and noisy sensor readings. 
\end{abstract}


\section{Introduction}
\label{sec:introduction}
\IEEEPARstart{T}{ypical} approaches to swarm robotics propose simple local behaviors for large numbers of simple robots such that they collectively accomplish a complex task; many approaches study the properties of the emergent behavior \cite{bayindir2016review,slavkov2018morphogenesis,kilobots}.
% \hkc{add citations}. 
% \sout{ by using stigmergic \hkc{always? you are not using that} communication and a series of simple local reactive behaviors.} 
In this work, we consider a swarm consisting of homogeneous 
% \sout{constrained}
% \hkc{what does "constrained" mean?}
robots which are
% which behave according to a simple set of rules, and the desired global behavior emerges from the interplay of local interactions.
% \hkc{this sentence is the same as the above} 
% The robots are 
minimalist; they have no memory, cannot broadcast or receive location information from their neighbors and are unable to plan ahead.
% \hkc{skipping the rest of this section and going to the contributions paragraph}
Minimalistic robotic swarms \cite{feola2022adaptive,selfPaper} have a number of applications, ranging from nanomedicine to underwater monitoring and surveillance \cite{babic2020novel,coquet2019local}, where robots might not be able to efficiently communicate with a central controller or with each other, and might not have the ability to self localize. For example, in an underwater mission, communication may be limited to acoustic signals, which are sensitive to interference and  lead to errors in the relative positioning of nearby entities. 

In this paper, we focus on the problem of encapsulating multiple targets, which are moving in unknown motion patterns, by a minimalist robotic swarm. A robot in the swarm has no knowledge of the exact relative location of nearby robots, targets, or the boundary of the environment. 
{This work extends our previous work on encapsulating static targets~\cite{selfPaper} by addressing moving targets. We develop an orbiting behavior for robots to encapsulate the targets, in addition to the searching for targets and avoiding collisions within the swarm, as in~\cite{selfPaper}. %  remains the same, we introduce an additional orbiting behavior for a robot to encapsulate a target in this paper. 
We compare the efficiency of our previous algorithm with the one introduced in this paper.
% We use our previous algorithm for target encapsulation as a baseline to compare the efficiency of the algorithm we introduce in this paper. 
Furthermore, we also show the behavior of our algorithm when applied to non-circular robots.  
% \hkc{cite and explicitly state what we added/changed. this needs to be in the intro}
}
\\

\noindent\textbf{Related Work:}  
There has been extensive work on developing various techniques to localize and track a moving target while ensuring collision avoidance \cite{xue2011swarm,hollinger2009efficient,li2022cooperative,hung2016scalable, DQ_catchMovingTarget,visualInterception}. 
%----- target tracking 
In \cite{Belkhouche}, authors introduced a motion planning strategy for a single robot based on velocity pursuit to intercept a target moving with unknown maneuvers. 
For target tracking using a multi-robot system, most approaches use artificial potential fields to design a controller consisting of a virtual attraction force to move towards a target and a repulsion force to avoid collision with obstacles \cite{xiong2009virtual}. 
%--- collision avoidance 
Another widely used approach to guarantee collision avoidance with dynamic obstacles is using a limit cycle method \cite{benzerrouk2012dynamic}.
% \hkc{add here that our approach is inspired by this work and how it is different}.
% In a first, \hkc{what does "in a first" mean?}
{The authors of \cite{hunting_limitCycle} introduced a hybrid approach where they instead used the limit cycle method to encircle a moving target using a swarm of holonomic robots, and artificial potential fields for collision avoidance. Since the use of the limit cycle method, either for surrounding a target or avoiding collision with obstacles requires the exact knowledge of the neighbor's relative position information, we cannot use it for our minimalist robotic swarm.  }
% \hkc{how is this paper different?}

Pursuit-evasion games~\cite{pe_1,pe_2,pe_3} provide guarantees for catching a faster-moving evader by constructing an encircling formation of pursuers composed of a series of Apollonius circles around a target and slowly closing the escape paths of the evader. In this approach, an evader is captured if a pursuer meets the evader at the same point at the same time. 
Most of the pursuit-evasion methods in the literature assume knowledge of the target's motion model.
% \hkc{the last 2 sentences say pretty much the same thing - do you need both?} 
In this work, we do not assume such knowledge.

Existing research \cite{wu2015hunting} in ``hunting" of dynamic targets generally makes use of communication within the team and formation-keeping control strategies, while approaching the target, to ensure that all of the escaping routes of the targets are occupied by the robots.
Work in \cite{hamed2020improvised} developed a leader-follower strategy based on the behavior of wolves to hunt a randomly moving target with unexpected behaviors. The authors of \cite{manzoor2017coordinated} proposed a limit cycle based algorithm using a neural oscillator to surround a target moving with unknown but constant velocity.
The authors of \cite{blazovics2012target} utilized rule-based mechanisms using only relative positions of neighbors and no direct communication within the swarm for surrounding an escaping target by introducing a circulating behavior in the swarm. 

Recent research in colloidal swarms has shown the capture of multiple randomly moving targets using self-organization control schemes. In \cite{xu2021brownian}, the authors designed a stochastic centralized controller for an intelligent colloidal micro-robotic swarm to capture multiple Brownian targets in a maze. In \cite{cargoCaptureSims}, the authors show via simulations, the feedback-controlled reconfigurability of colloidal particles that act as a swarm capable of capturing and transporting microscopic Brownian cargo. 

{To implement a distributed approach of searching and encircling targets in an inexpensive and efficient way,} in \cite{lee2010tracking} the authors developed a new dual-rotating proximity sensor to obtain relative position information of neighbors for tracking multiple targets with a minimalist swarm. 
Authors of \cite{franchi2016decentralized} proposed a scheme to estimate the global quantities required by the controller in a decentralized way using only local information exchange between robots for the guaranteed encirclement of a 2D or 3D target. 
% \hkc{what is the purpose of this paragraph? looks like it describes both control and sensing. how is this paper related to these works?}

While the above approaches successfully solve the target encirclement while avoiding collisions, most of them rely on the assumption that robots have knowledge of the exact relative location of both their neighbors and the target. Furthermore, it is a common assumption that the average speed of the agents in the swarm is greater than that of the moving target to guarantee encapsulation \cite{lumelsky1997decentralized}. 
In contrast, in this work, we provide guarantees on the encapsulation of dynamic targets without the requirement of accurate (relative) location information and without direct communication within the swarm. % which is an interesting problem because of its real-world applications in constrained environments. 
\newline
% This work extends our previous paper \hkc{cite and explicitly state what we added/changed. this needs to be in the intro}
\newline
\noindent\textbf{Contributions:} This paper's contributions are: (i) a discrete-time decentralized control law for a minimalist robotic swarm that guarantees the encapsulation of dynamic targets, {for different target motion models,} without accurate detection of the relative location of either the targets or neighboring robots, given certain bounds are met (ii) {sensor-placement} dependent bounds on the ratio between the target and robot speeds to guarantee encapsulation, (iii) proof of stochastic convergence of our control law for scenarios when a target is moving faster than a robot, and (iv) simulations and analysis of emergent behavior of the swarm in the presence of sensor noise and different task parameters.

%%%%%%%%%%%%%%%%%%%%%%%%%%%%%%%%%%%%%%%%%%%%%%%%%%%%%%%%%%%%%%%%%%%%%%%%
\section{Definitions}
In this section, we provide definitions from~\cite{selfPaper} that we use throughout. 
\\

% \subsection{Environment, and Robot Model}
\noindent\textbf{Environment:} We consider a $2D$ convex bounded
% \hkc{does it have to be convex and bounded? do we even need a boundary?} 
environment $E \subseteq \mathbb{R}^2$. The environment has a fixed global frame {$\mathcal{I}$}. \\
% \newline
% \noindent\textbf{Target:} 
% \hkc{i suggest moving this to the problem formulation. This is not the same as in the iros paper, as opposed to the environment and robot model that stay the same} 

\noindent\textbf{Robot:} We model a robot, $R = (\mathbf{c}_r,\gamma_r,r_{r},p,Z)$,
% \hkc{why did you change gamma to theta?} 
as a disk of radius $r_{r}$ centered at $\mathbf{c}_r \in E$ with heading $\gamma_r \in \mathbb{S}$. 
% as shown in Fig. \ref{fig:robModel}
% \begin{figure}[h]
%   \centering
%   \includegraphics[width=0.48\linewidth]{figures/robotModel.png}
%   \caption{Robot model with a total of 5 sensors arranged asymmetrically on its periphery.}
%   \label{fig:robModel}
%   \Description{Robot model with a total of 5 sensors arranged asymmetrically on its periphery.}
%     \vspace{-1.3em}
% \end{figure}
The shape of a robot does not affect the analysis presented in the paper since the robot can always be circumscribed by a circle of radius $r_r$. 
Each robot is reactive, memoryless, has no knowledge of the relative locations of other robots or targets, and cannot communicate with its neighbors.
The kinematics of a robot is given by Eq.~\eqref{robotModel}, which is a typical model for a differential drive robot. At each time step, the robot is controlled in a \textit{turn-then-move} scheme 
with control inputs $\theta_r \in \mathbb{S}$ and $d_r \in \mathbb{R}^+$. The maximum step-size of a robot is $d_r^\mt{max}$. 
% \hkc{ok, looks like you flipped the theta and gamma with respect to the iros paper. was that on purpose? i would suggest being consistent with the iros paper}
%--Change wording, IROS paper
% At each time step, $\theta_r \in \mathbb{S}$ and $d_r \in \mathbb{R}^+$ are the control inputs corresponding to the angle turned and the distance moved by a robot. 
% The maximum distance a robot can move at each time step is limited to $d_r^\mt{max}$. 
\begin{align} \label{robotModel} 
    \gamma_{r,T} &= \gamma_{r,T-1} + \theta_r \nonumber \\
    \vb{c}_{r,T} &= \vb{c}_{r,T-1} + d_r[\mt{cos}\gamma_{r,T} \quad \mt{sin}\gamma_{r,T}]
    % \vspace{-2em}
\end{align}
A robot has $p$ isotropic sensors arranged on its boundary such that $\phi^k \textrm{ }\forall k\in \{1 \cdots p\}$ is the angle between the $k^{th}$ sensor and the robot's heading direction. $Z$ is the set of measurements from all sensors on a robot.
\\

\noindent\textbf{Signal Sources:}
We consider three types of signal-emitting sources present in the environment that a robot can detect: $s_g$ from a point source at the center of a target, $s_r$ from a point source at the center of a robot, and $s_e$ from a line source present on the entire environment boundary. For clarity in notation, we hereby denote the signal set $\{s_g,s_r,s_e\}$ by $\{g,r,e\}$. 
% of type $s_g$ by subscript $g$, $s_r$ by subscript $r$, and $s_e$ by subscript $e$.

The strength of any signal $s \in \{g,r,e\}$ located at a distance $d$ from a signal source is given by the function $B_s(d)$.  
The influence distance of a source is limited to $\beta_s$, such that $B_s(d) = 0 \textrm{ }\mt{ } \forall d \geq \beta_s$. 
% \amy{types of signal (after reading the rest of the paragraph, I'm not actually sure what you're referring to here as a source)} 
% types of signals. 
Let $N^k_s$ be the set of all the sources of type $s$ in the sensing range of the $k^{\mt{th}}$ sensor and $d_j^k$ be the distance of this sensor from a source $j \in N^k_s$.  
Then the sensor reading $z_s^k = \sum_{j \in N^k_s} B_s(d^{k}_j)$ is the sum of signal strengths from all sources in $N^k_s$.
% as defined in Eq. (\ref{totalIntesnity}). 
% \begin{equation} 
% \label{totalIntesnity}    
%     z_s^k = \sum_{j \in \mathcal{N}^k_s} B_s(d^{k}_j)
% \end{equation}
This summation becomes an integral over the boundary segment for a line source present inside the influence region $\beta_e$.

The tuple $(z_g^k,z_r^k,z_e^k)$ corresponds to the measurements of the $k^{th}$ sensor. 
Let $Z_g = \{z_g^1 \cdots z_g^p\}$, $Z_r = \{z_r^1 \cdots z_r^p\}$ and $Z_e = \{z_e^1 \cdots z_e^p\}$, then the measurement set is $Z = Z_g \cup Z_r \cup Z_e$.
We define $r^\mt{safe}_s \textrm{ }\forall s \in \{g,r,e\}$ as the user-specified minimum safety distance that a robot must maintain from a source at all times.

\section{Problem Formulation}
\label{section_problemForm}
We model a target $g =(\mathbf{c}_{g},r_{g})$ as a disk of radius $r_{g}$ centered at $\mathbf{c}_{g}\in E$. 
% \amy{what about when c is in E but the outer part of the disk is not in E?}. 
$\mathcal{G}$ is the set of all targets contained in $E$. The kinematics of a target is given in Eq.~\eqref{targetModel}. At any time step $T$, $d_{g} \in \mathbb{R}^+$ is the distance moved by the target, and $\gamma_{g,T} \in \mathbb{S}$ is the target heading.
% \amy{I don't get what the difference between $\gamma_{g,T}$ and $\gamma_{g}$ is}
\begin{align} \label{targetModel} 
    \gamma_{g,T} &= \gamma_{g,T-1} + \theta_g \nonumber \\
    \vb{c}_{g,T} &= \vb{c}_{g,T-1} + d_{g}[\mt{cos}\gamma_{g,T} \quad \mt{sin}\gamma_{g,T}]
\end{align}
The maximum distance that a target can move is $d_g^\mt{max}$. 
\newline
\newline
\noindent\textbf{Target Motion Models:}
\label{section_targetMotion}
% Let $r_{ji}$ be the relative distance between a target $g \in \mathcal{G}$ and $i^{\mt{th}}$ robot. 
In this paper, we design controllers and analyze the swarm behavior for different types of target motion models. A target can exhibit one of the following motions:
\begin{enumerate}
    \item \label{randMotion} Target moves randomly such that at any time step $T$, $\gamma_{g,T} \in [0\textrm{ } 2\pi)$, $d_{g} \in [0 \textrm{ } d_g^{max}]$ and $\vb{c}_{g,T} \in E$.
    \item \label{randMotionEscape} Target moves randomly as in motion model~\ref{randMotion} until a robot is in its \textit{escape domain} = $(\vb{c}_g,r_g^{\mt{escape}})$ of radius $r_g^{\mt{escape}}$ centered at $\vb{c}_g$, in which case the target chooses a heading direction to escape from all the robots {that satisfies}  $\norm{\vb{c}_{g,T}-\vb{c}_{r,T}} \leq r_g^{\mt{escape}}$, and moves the maximum step-size $d_{g}$.
    \item \label{predMotion} Target follows an unknown motion pattern until a robot satisfies $\norm{\vb{c}_{g,T}-\vb{c}_{r,T}} \leq r_g^{\mt{escape}}$, in which case it chooses a heading direction to escape nearby robots. % \amy{i'm confused, this is saying that the target is actively escaping the robots but in an unknown way?}
    \\
\end{enumerate}

\begin{figure}[!ht]
  \centering
  \includegraphics[width=\linewidth]{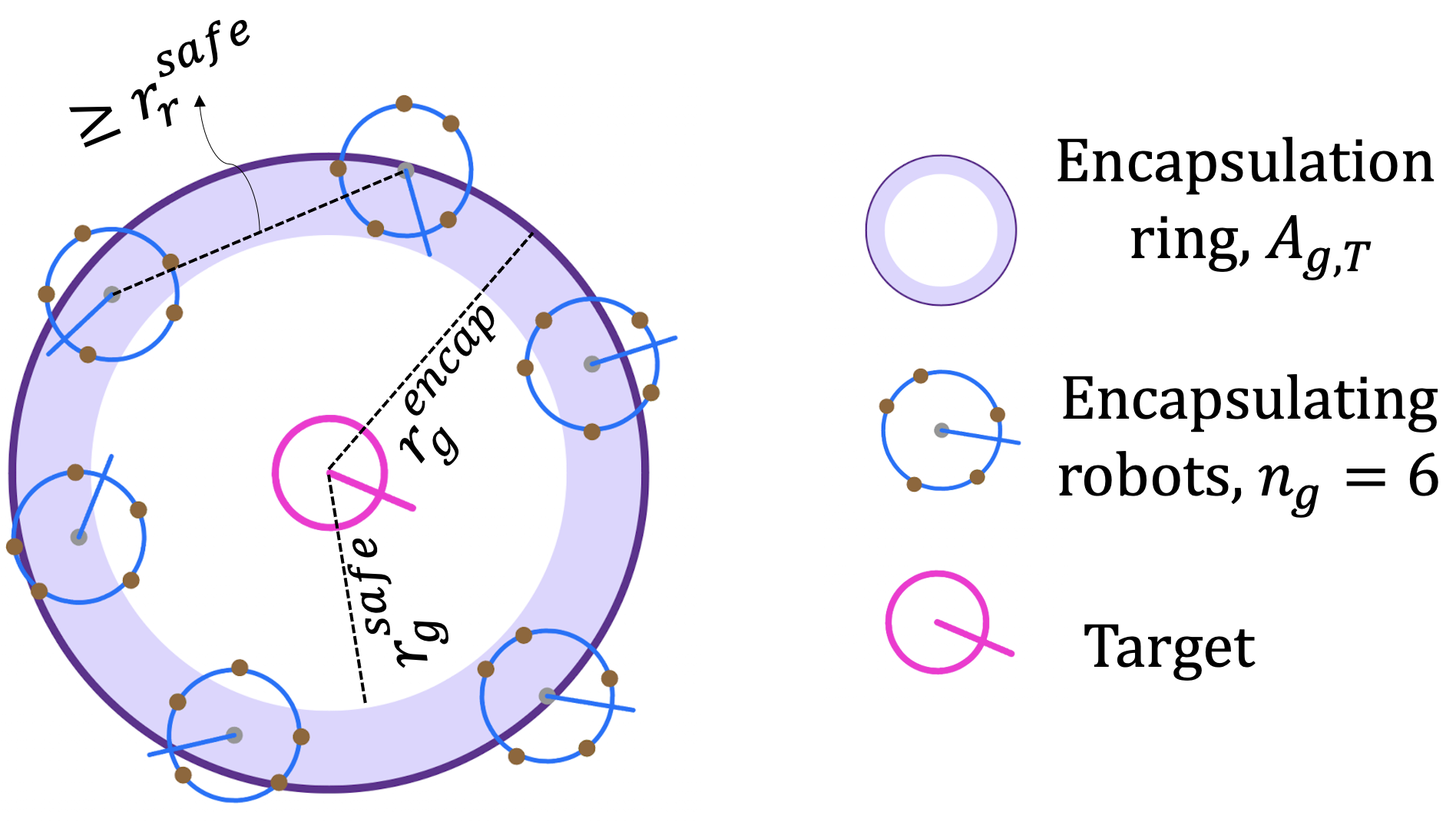}
  \caption{A target is encapsulated if $n_g$ robots are present simultaneously in the encapsulation ring while maintaining at least a distance of $r_r^{\mt{safe}}$ from each other.}
  \label{fig:encap}
\end{figure}

\noindent\textbf{Target Encapsulation:} For each target $g \in \mathcal{G}$, we define an encapsulation ring $\mathcal{A}_{g,T} = (\mathbf{c}_{g,T},r^\mt{safe}_g,r^\mt{encap}_g)$ of inner radius $r^\mt{safe}_g$ and outer radius $r^\mt{encap}_g$ centered at $\mathbf{c}_{g,T}$. 
A robot $R$
% \hkc{should this be r? not R?} \himani{I defined in Prelim section, a robot to be denoted by R} 
is considered to be in $\mathcal{A}_{g,T}$ if, $r^\mt{safe}_g<\norm{\vb{c}_{r,T}-\vb{c}_{g,T}}\leq r^\mt{encap}_g$.
% \begin{equation} \label{trappingCondition}
%     r^\mt{safe}_g<\norm{\vb{c}_{r,T}-\vb{c}_{g,T}}\leq r^\mt{encap}_g
% \end{equation}
A target is \textit{encapsulated} if the total number of robots present in the encapsulation ring is $n_g$, which is a user-specified input as shown in Fig.~\ref{fig:encap}.
% \hkc{add image}
\\

% \subsection{Problem Formulation}
\noindent\textbf{Problem statement:}
Consider a bounded environment $E \subseteq \mathbb{R}^2$ with $m$ dynamic targets where the initial distribution of the robots and targets is arbitrary. 
Given the total number of sensors $p$ on a robot, the user-provided safe distance $r^{\mt{safe}}_s$ $\forall s \in \{g,r,e\}$, the encapsulation ring $\mathcal{A}_g$, and the number of robots $n_g$ needed to encapsulate each target $g$ such that the total number of robots $n \geq \sum_{g \in \mathcal{G}} n_g$, our objective is to find a real-time decentralized control law for encapsulating all targets while ensuring safety distances are always maintained. %Each robot operates autonomously in the environment and they do not share a common coordinate frame or broadcast messages within the swarm.
% \noindent \textbf{Assumptions:}
% \noindent 
We make the following assumptions about the environment and the system:
\begin{assum}
\label{sensorPlacement}
The sensors are arranged on a robot such that when a robot's center is $r_s^{\mt{safe}}$ away from a source $s$, at least one sensor is in the influence region of the source. 
For ease of exposition, we consider circular robots with a symmetric placement of sensors to explain our algorithm, and show in simulations how asymmetric sensor placements and non-circular robots affect swarm behavior. 
% \hkc{weren't you going to simulate asymmetric placements?}
%---
\end{assum}

\begin{assum}
\label{only1TarSensed}
The distance between any two moving targets is greater than ($2\beta_g+2r_r$). That is, a robot can sense at most one target at a time. 
% \amy{i would reframe this so that the second sentence comes before the first}
\end{assum}

\begin{assum}
\label{targetMotionNearEnv}
We constrain a target to maintain a minimum distance of ($r_g^{\mt{encap}} + r_e^{\mt{safe}} + d_r^{\mt{max}}$) from the environment boundary. This ensures that robots will be able to encapsulate the target without colliding with the environment boundary. 
\end{assum}

\begin{assum}
\label{targetAssum}
We place no restriction on the target's knowledge of the environment; it may be able to perfectly sense the relative location of any robot present in its user-specified \textit{escape domain}, $r_g^{\mt{escape}}$ thereby knowing the optimal escape route.  
% \hkc{why do you need the following assumption? not the previous, the following}
% \himani{to give a better view of the algorithm: that is, a minimalist swarm able to catch an intelligent target (knows perfect location). We can have a target not able to perfectly locate, but that will benefit swarm.}
However, if a target is encapsulated, we assume it emits a single burst of a shut-off signal and stops emitting any signal subsequently.
The influence distance of this signal is limited to $\mathcal{A}_g$, and we assume that thereafter both the robots within the encapsulation ring and the target stop moving, i.e. $d_r = 0$ and $d_g = 0$, respectively. 
\end{assum}

\begin{assum}
\label{inverseBKnown}
The signal strength $B_s$ strictly decreases with the radial distance, $d$ from a source and the inverse of the signal function $B_s(d)$ exists and is known to the robots. 
\end{assum}

%%%%%%%%%%%%%%%%%%%%%%%%%%%%%%%%%%%%%%%%%%%%%%%%%%%%%%%%%%%%%%%%%%%%%%%%
\section{Approach}
% Given the strict constraints on a swarm robot's capability,
Our strategy for designing a local control law is based on geometry and the relative kinematics of the interaction of a robot with its neighboring robots and a dynamic target. We extend our previous work \cite{selfPaper} where we only considered static targets; a robot's behavior there was to either move randomly in the bounded environment when it does not sense any target, or to move towards a target if sensing one while ensuring safety. 
Here, we introduce an additional robot behavior of orbital encirclement of a target, inspired by \cite{benzerrouk2012dynamic}. As we show in Section~\ref{sec_proof}, this behavior ensures the encapsulation of an escaping target. 
In Section~\ref{section_vs} we describe virtual sources as defined in \cite{selfPaper} and use them to under-approximate the relative distance between a source and the robot's center as a function of the sensor placement. In Section~\ref{section_ca} we find the bounds on control parameters ($d_r$ and $\theta_r$) for a robot to ensure that it maintains $r_s^{\mt{safe}}$ distance from a source $s \in \{g,r,e\}$. 
In Section~\ref{section_orbits} we introduce the concept of orbital encirclement of a moving target; we provide a summary of the overall reactive control law for a robot in the swarm in Section~\ref{section_controlSummary}.

\subsection{Virtual Source}
\label{section_vs}
\begin{figure}[!ht]
  \centering
  \includegraphics[width=\linewidth]{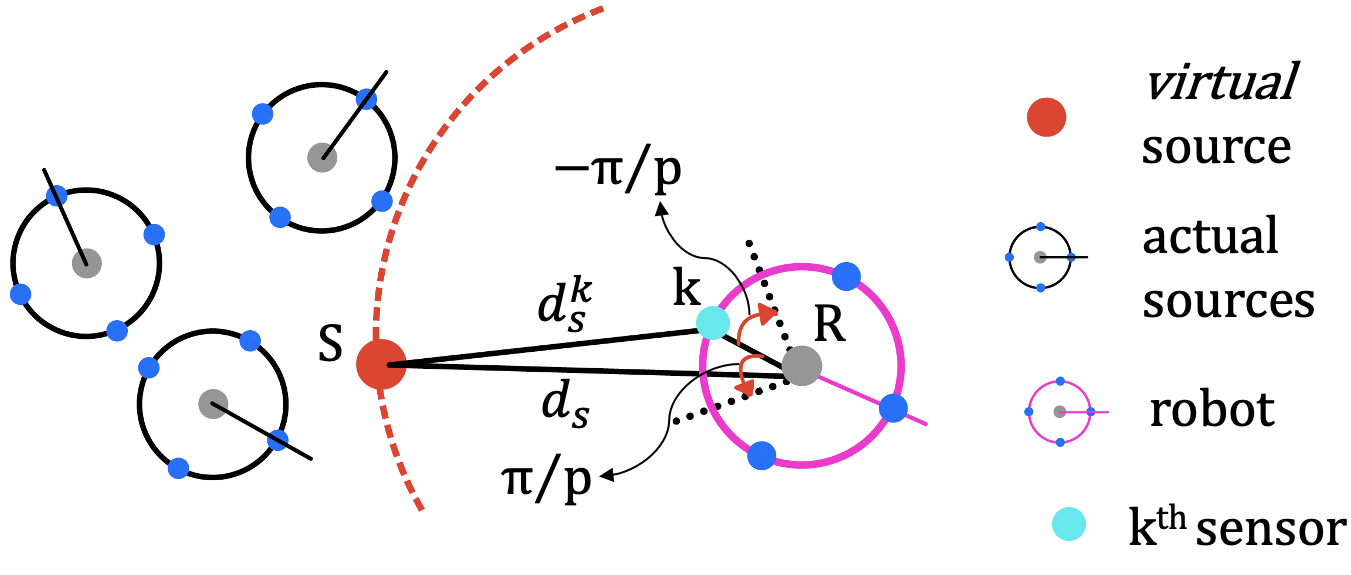}
  \caption{Virtual source for the $k^{\mt{th}}$ sensor ~\cite{selfPaper}. 
  % \hkc{the sensor is labeled in the image with A but we are refering to the kth sensor, this is confusing}
  }
  \label{fig:vs_iros}
\end{figure}

Since we assume a robot is equipped with isotropic sensors, a sensor measurement corresponds to the aggregated signal strength from all the nearby sources. Hence, the same measurement could correspond to a single source nearby or a cluster of sources further away. Therefore,
% we use virtual source as defined in \cite{selfPaper} to estimate the location of nearby entities.
% As outlined in \cite{selfPaper}, 
for each sensor reading, $z_s^k$ $\forall s \in \{g,r,e\}$, we define a virtual source on a circle centered at the sensor $k$ as shown in Fig.~\ref{fig:vs_iros}.
% and use the Assumption ~\eqref{inverseBKnown} to find its radial location, $d_s^k = B^-_s(z_s^k)$. 
% Let $k^{\mt{th}}$ sensor be receiving the strongest signal from actual sources of type $s$, i.e. $z^k_s \geq z^l_s, \forall l\neq k$. 
It is shown in \cite{selfPaper} that the closest possible location of the virtual source with respect to the robot's center is given by Eq. ~\eqref{dsk}. Furthermore, the range of possible directions of the location of the virtual source with respect to the robot's center is restricted to $[\phi^k-\pi/p,\textrm{ }\phi^k + \pi/p]$ for symmetric sensor placement. 
\begin{equation} 
    \label{dsk}
   d_s = r_r\mt{cos}(\pi/p) + \sqrt{(d_s^{k})^2-r_r^2\mt{sin}^2(\pi/p)}
\end{equation}
For asymmetric sensor placement, we replace $\pi/p$ with half of the maximum angle that the $k^{\mt{th}}$ sensor makes with either of its adjacent sensors. Similarly, for robots that are not circular in shape, we replace $r_r$ by the distance between the $k^{\mt{th}}$ sensor and the robot's center in the above equation.
We can see in ~\eqref{dsk} that as $p \rightarrow \infty, d_s \rightarrow d_s^k + r_r$. That is, the error in locating the source is dependent on the total sensors on a robot. 
% \hkc{algo 1 is very far from here so i do not think it makes sense to put this here} 
% \hkc{why did you leave this here? it is too far from the algo}
% In Algorithm \ref{algo:ca1}, the variables \textbf{\texttt{DistToTar}} and \textbf{\texttt{DistToEnvBound}} corresponding to a robot's estimates of the distances to a target and the environment boundary are computed using Eq. (\ref{dsk}) and the sensor receiving maximum intensity from the target and environment boundary, respectively.

\subsection{Collision Avoidance}
\label{section_ca}
% \hkc{this section should be in the approach. this paper is a superset of the iros paper + the extra work. we need that info in here as well, assuming we can fit it. you can refer the proofs to the iros paper, but this is not a definition}
% - with both static and dynamic obstacle \\
% - brief introduction with reference to IROS Paper
We use the technique introduced in \cite{selfPaper} for collision avoidance with nearby robots and the environment boundary. 
At each time step, the robot estimates the relative distance between its center and the nearby sources using Eq. ~\eqref{dsk} for the sensor with the maximum sensor reading $z_s^k$ $\forall s \in \{g,r,e\}$. 
If this distance is less than or equal to $(r_s^{\mt{safe}} + d_s^{\mt{max}})$, 
% \hkc{i do not think $d_r^{\mt{safe}}$ was introduced yet} 
the collision avoidance behavior is triggered for this robot to ensure safety.

\begin{figure}[!ht]
  \centering
  \includegraphics[width=\linewidth]{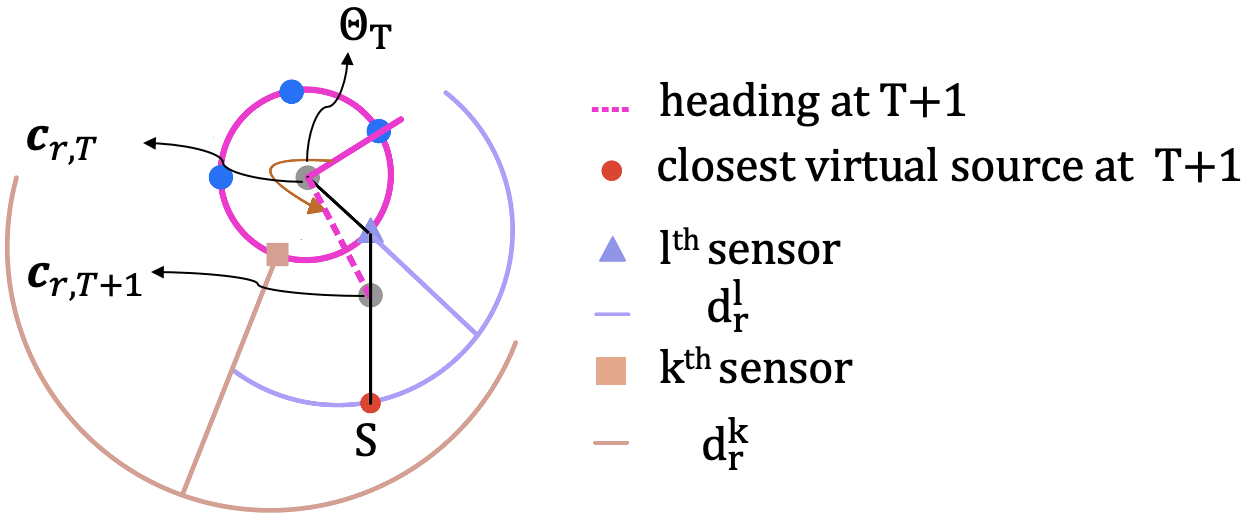}
  \caption{Computing $d_r$ such that collision is avoided with nearby moving robots  ~\cite{selfPaper}.
  % \hkc{again, there is an 'A' in the image. should closest source be "closest virtual source"? otherwise it is confusing as to why the k radius is bigger}
  }
  \label{fig:dynColl}
\end{figure}

We have shown in \cite{selfPaper} that to avoid collisions with static obstacles (such as environment boundary), the robot's heading direction $\theta_r$ must be chosen from the angular range given by Eq. ~\eqref{avoStaticObs}.
% \begin{figure}[h]
%   \centering
%   \includegraphics[width=0.75\linewidth]{figures/tarAtt_iros.png}
%   \caption{$\Theta_e^{\mt{avo}}$ ~\cite{selfPaper}.}
%   \label{fig:vs_iros}
% \end{figure}
\begin{equation}
    \label{avoStaticObs}
    \Theta_e^{\mt{avo}} = [\phi^k + \pi/p + \pi/2, \textrm{ } \phi^k - \pi/p + 3\pi/2] 
\end{equation}
Whereas to avoid the neighboring moving robots, the distance $d_{r}$ that a robot moves at time step $T$ in a given heading direction $\gamma_{r, T}$ must be chosen such that at $T+1$ it maintains at least a distance of $r_r^{\mt{safe}}$ from the closest neighboring robot. 
As shown in Fig.~\ref{fig:dynColl}, let $k$ and $l$ be the indices of the sensors closest to the intended heading direction $\gamma_{r}$ at time $T$ and $d_r^k$ and $d_r^l$ are their radii of virtual sources respectively such that $d_r^k > d_r^l$. Then, we can compute the bounds on the step-size $d_{r}$ that the robot can take in the heading direction $\gamma_{r,T}$ using Eq. ~\eqref{avoObsGivenTh}. 
% \cite{selfPaper}. 
\begin{multline}
    \label{avoObsGivenTh}
    0 \leq d_r \leq r_r \mt{cos}(\phi^l-\theta_r) \\
    + \sqrt{(d_r^l - r_r^{\mt{safe}} - d_r^{\mt{max}})^2 - r_r^2 \mt{sin}^2(\phi^l-\theta_r)}
\end{multline}
% \hkc{same comment about algo 1} 
% \hkc{why did you leave this here? it is too far from the algo}
% In Algorithm \ref{algo:ca1}, the function \textbf{\texttt{DistAvoRob}} computes the maximum possible value of $d_r$ from Eq. ~\eqref{avoObsGivenTh}.

To ensure that two robots never deadlock, the bounds on the maximum step size a robot can take, and the influence region of a robot's source, are given by Eq. ~\eqref{bound_d_r_max} and Eq. ~\eqref{bound_beta_r}, respectively. The proof is detailed in Lemma V.3 of ~\cite{selfPaper}.
% \cite{selfPaper}.
\begin{multline}
    \label{bound_d_r_max}
    d_r^{\mt{max}} < \frac{r_r^{\mt{safe}}+r_r\mt{cos}(\pi/p)}{2} \\
    -\frac{\sqrt{(r_r^\mt{safe})^2 + r_r^2 - 2r_rr_r^\mt{safe}\mt{cos}(\pi/p)}}{2}
\end{multline}
\begin{multline}
    \label{bound_beta_r}
    \sqrt{(r_r^\mt{safe})^2 + r_r^2 - 2r_rr_r^\mt{safe}\mt{cos}(\pi/p)} + 2d_r^{\mt{max}} < \beta_r \\
    < r_r^{\mt{safe}}+r_r\mt{cos}(\pi/p)
\end{multline}

\subsection{Encirclement of a Target}
\label{section_orbits}
{In \cite{selfPaper}, our approach to encapsulate a static target, was for a robot to either move towards the target or move away from an obstacle between itself and the target in the direction of the sensor receiving the minimum reading from nearby moving robots.} 
%- talk about the previous approach for a static target
%- why would it not work for dynamic
%- current algo can be used for both static and dynamic target
% \hkc{you need an overview of this before delving in. maybe talk about the other paper?}
However, in order to surround a dynamic target, the behavior of a robot should be such that the swarm is able to disperse around the target in order to block off its escaping paths. 
Since we consider minimalist robots that can neither communicate with their neighbors nor know their exact relative position, we can not make use of formation control strategies, such as \cite{manzoor2017coordinated,hunting_limitCycle}.
% \hkc{add citations}

Consider a scenario where all the robots in the swarm start on one side of a target. {Then, for a swarm to disperse around a target, it is necessary that an individual robot be able to catch up with the escaping target, and once the robot reaches the encapsulation ring, it should be able to encircle the target so that the target is prevented from escaping.}
% \change{Then, it is necessary that robots in the swarm be able to catch up with the escaping target and shut all of its escaping directions.} 
% That is, a necessary condition to guarantee the encapsulation of a target is that an individual robot in the swarm has the ability to complete a revolution around the target.
% \hkc{still not clear how the last sentence is a consequence of the sentence before it. if anything, i was expecting "the robot will be able to outrun the target"}
% \hkc{why?what if it can only do half a perimeter?}
% instead of being huddled together on one side of the target when it is in escape mode
% \amy{this sentence is confusing}.
% To find the control parameters for a scenario when a robot is inside the influence region of a target
% and ensure that the robots are able to encircle an escaping target 

\begin{figure}[!ht]
  \centering
  \includegraphics[width=\linewidth]{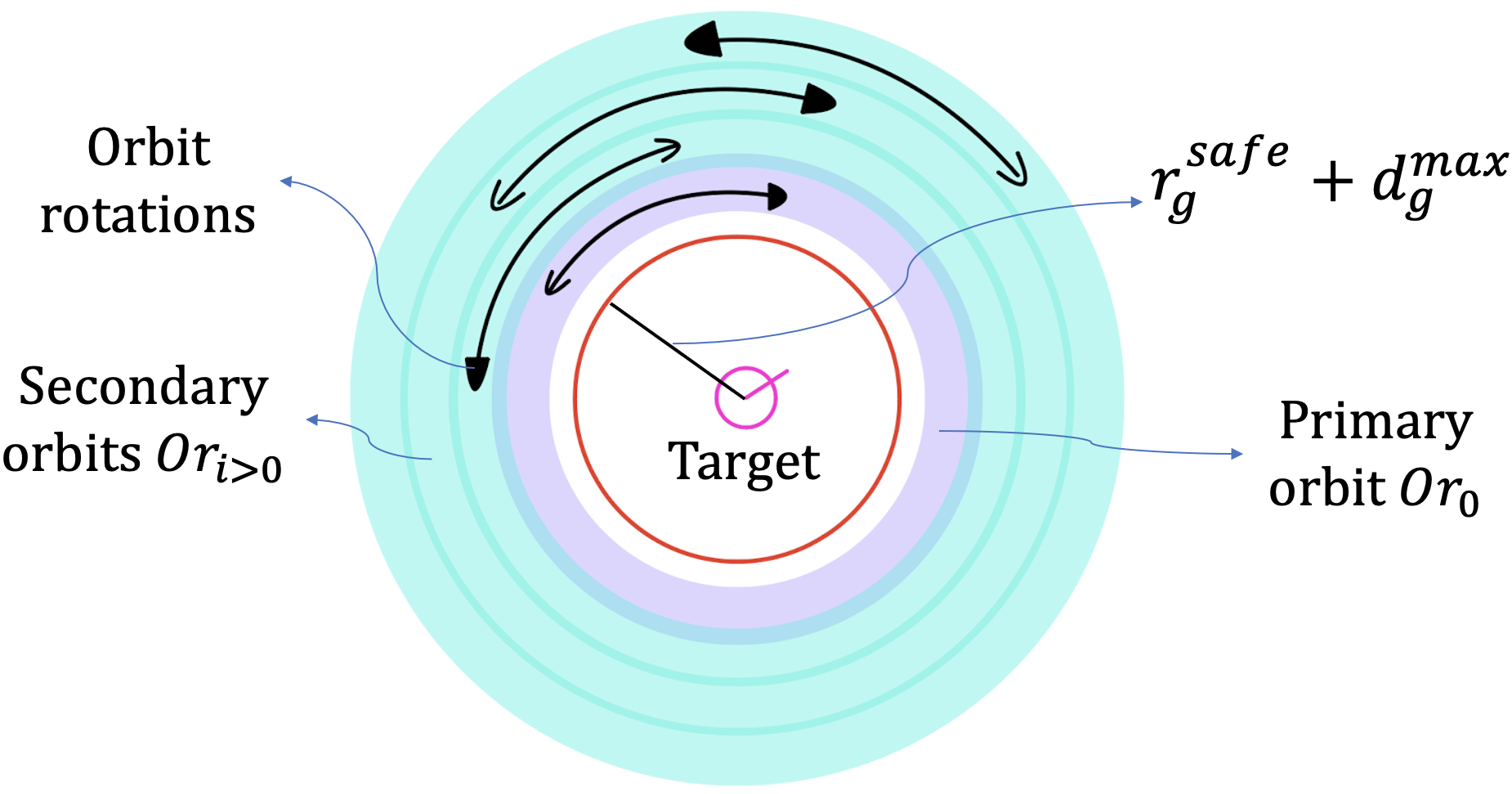}
    % \vspace{-1.3em}
  \caption{Primary (purple ring) and secondary (cyan rings) orbits around a target, and the lower bound on the target's escape domain (red circle) . A robot moves either clockwise or counter-clockwise in an orbit depending on its neighbors. The solid arrows denote the tie-breaking rotation for an orbit.
  % \hkc{i would remove, from the figure, "the escape domain of the target" - it is confusing because it looks like the escape domain is exactly what is in the red circle, but that is just a lower bound. When we talked we said you should write "minimal" but now i think you should just remove.} %\hkc{i cannot tell which side it the "solid arrow" - might be worth making the arrow triangle bigger}
  % \hkc{why are the arrows double sided? looks like each orbit can go in both directions. ok, after reading further, i get it but then it needs to be explained that it is a tie breaker before}
%\hkc{might be good to also show the escape region on this figure}  
}
  \label{fig:orbit}
    % \vspace{-1.3em}
\end{figure}

To ensure encapsulation, we define primary and secondary orbits around each target, as shown in Fig.~\ref{fig:orbit}.
% it is currently sensing. 
For each orbit, we define a \textit{tie-breaking orbital rotation} which can be either clockwise (denoted by a value of -1) or counter-clockwise (denoted by a value of 1). 
% Intuitively, 
% We need to make sure that the robots are able to disperse around and not huddle together on one side of the target when it is in escape mode.
The primary orbit, $Or_0 = (\vb{c}_{g,T},Or_0^{\mt{inner}},Or_0^{\mt{outer}},-1)$ 
% \hkc{why '{}' and not '()'? }
is an annular ring centered at $\vb{c}_{g,T}$ with an inner radius of $Or_0^{\mt{inner}} \geq r_g^{\mt{safe}} + d_g^{\mt{max}}$, an outer radius $Or_0^{\mt{outer}} = r_g^{\mt{encap}}$ and a clockwise orbital rotation (chosen arbitrarily). 
Let $w$ be the width of a secondary orbit, then
an $i^{\mt{th}}$ secondary orbit is given by, 
$Or_i = (\vb{c}_{g,T},Or_0^{\mt{outer}}+(i-1)w,Or_0^{\mt{outer}}+(i)w,(-1)^{i-1}), \textrm{ }\forall i > 0$.
% \hkc{shouldn't this be $(-1)^{i-1}$?}
% The total number of orbits for a target are such that its entire influence region (of radius $\beta_g$) is covered. 
We consider a robot to be in $i^{\mt{th}}$ orbit if,
$Or^{\mt{inner}}_i<\norm{\vb{c}_{g,T}-\vb{c}_{r,T}} \leq Or_i^{\mt{outer}}$.
Each robot in the swarm computes its \textit{current orbit} using its estimate of $\norm{\textbf{c}_{g,T} - \textbf{c}_{r,T}}$.
At time-step $T$, let $Or_i$ be the current orbit as estimated by a robot, then its control consists of one of the following behaviors:
% \hkc{is this a prioritized list? i got to here}:
\begin{enumerate}
    \item \textit{if} $i > 0$, the robot moves towards the target in a heading direction chosen  from the line of sight angular range as estimated from the virtual source
    $(\Theta_g^{\mt{LOS}})$ 
    while maintaining a safe distance from nearby robots.
    % of $r_r^{\mt{safe}}$ from the robots in the front. 
    \item \textit{else if} $i > 0$ and the robot cannot move a non-zero distance towards the target, it moves tangentially in its current orbit while maintaining a safe distance from nearby robots.
    % and $r_g^{\mt{safe}}$ from the target. 
    The direction of the tangent
    % (clockwise or anti-clockwise) 
    is chosen such that it maximizes the possible step-size $d_r$. In case of symmetry,
    % in the possible step-size $d_r$
    the robot moves in the \textit{orbital rotation} of the $i^{\mt{th}}$ orbit.
    \item \textit{else if} $i > 0$ and the robot can neither move in a direction from  $\Theta_g^{\mt{LOS}}$ nor tangential to the orbit, it chooses a direction of motion that maximizes the possible step-size $d_r$.
    % \item if $i = 0$ and the robot cannot move tangential to the orbit, it moves away from the target. 
    % The angular range for this motion is similar to the one given by Eq. ~\eqref{avoStaticObs}
    \item \textit{else if} $i = 0$, the robot moves tangentially in its current orbit while maintaining a safe distance from the target.
    \item \textit{else if} the relative distance between the target and a robot is less than or equal to $Or_0^{\mt{inner}}$, it moves away from the target.
    \item \textit{else} the robot performs a simple random walk while avoiding nearby moving robots.
\end{enumerate}
In general, a robot moves toward the target until it reaches the primary orbit. If other robots are present between itself and the target, the robot moves tangentially in its current orbit until it can move toward the target. All the robots that place themselves in the primary orbit constantly move tangentially and eventually close off the target's escape routes. % for the target. 
The width of a secondary orbit, $w$, must be less than $\beta_{r}$, so that a robot's neighbors in adjacent orbits lie within its sensing range. This ensures that a robot doesn't move towards a target when it senses other robots in the front and instead moves tangentially in its current orbit. 

\begin{figure}[!ht]
  \centering
  \includegraphics[width=0.9\linewidth]{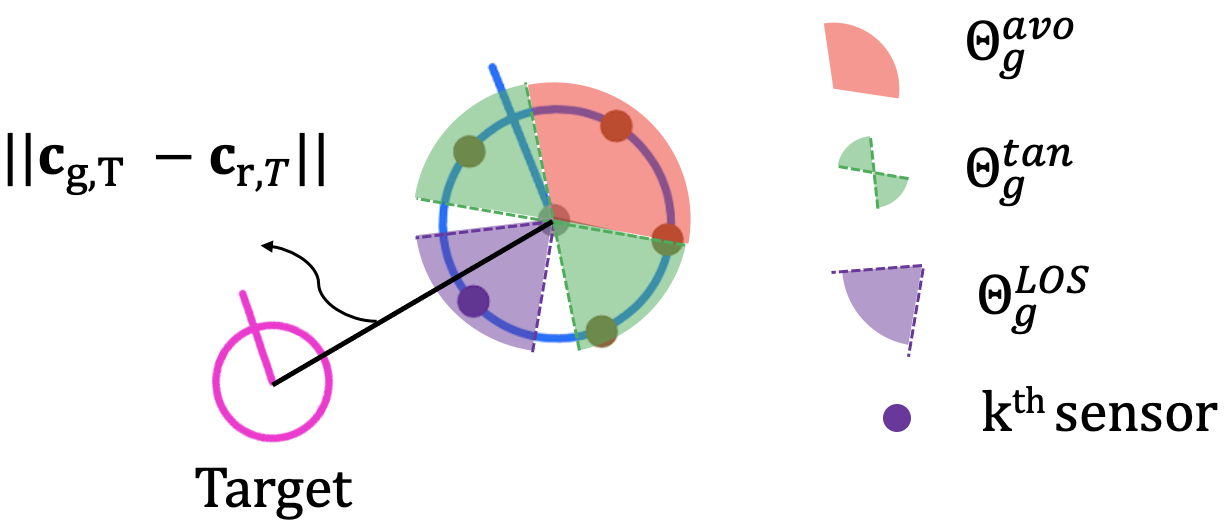}
    % \vspace{-1em}
  \caption{The angular range set for a target-robot interaction. The robot is equipped with 5 sensors placed asymmetrically.  
%   \amy{shouldn't the black line be in the exact middle of $\Theta_g^{\mt{LOS}}$ }
  % \amy{Maybe I missed this - why is the theta\_LOS not symmetric about the sensor?}
  % \vspace{-1.3em}
  }
  \label{fig:angRangesTar}
\end{figure}

Now, using the sensor readings and their corresponding virtual sources, we find the set of directions that a robot needs to choose from to move towards a target, away from a target, or tangentially in an orbit. 
% Since a robot can only sense one target at any time $T$ (Assumption \ref{only1TarSensed}), 
Let $k$ be the index of the sensor such that $z^k_g > z^l_g, \forall l\neq k$. 
Here we have ignored the unlikely
% \amy{i don't get what you mean by "ignored"}
% the perfectly symmetric 
scenario where two sensors receive the same maximum intensity from a target.
% \amy{why?}. 
% Using virtual source defined in Section \ref{section_vs}, 
Then the angular range, $\Theta_g^{\mt{LOS}}$, for the possible location of the target with respect to the robot's center is given by Eq. ~\eqref{interval_of_tar}.
% \amy{grammar - not sure what you're trying to say}.
\begin{equation}
    \label{interval_of_tar}
    \Theta_g^{\mt{LOS}} = [\phi^k-\pi/p,\textrm{ }\phi^k + \pi/p]
\end{equation}
% \amy{i don't think you ever defined $\Theta_g^{\mt{LOS}}$ before this}
The angular range, $\Theta_g^{\mt{avo}}$ (Eq. ~\eqref{interval_avo_tar}), to move away from the target can be derived in a similar fashion to Eq. ~\eqref{avoStaticObs}.
% as in Section \ref{section_ca}.
\begin{equation}
    \label{interval_avo_tar}
    \Theta_g^{\mt{avo}} = [\phi^k + \pi/p + \pi/2, \textrm{ } \phi^k - \pi/p + 3\pi/2] 
\end{equation}
The angular range to move tangentially in an orbit in a clockwise or counter-clockwise direction is given by Eq. ~\eqref{interval_tan_ckw} and Eq. ~\eqref{interval_tan_antiCkw}, respectively, where we define $\Theta_g^{\mt{tan}} = \Theta_g^{\mt{tan,+1}} \cup \Theta_g^{\mt{tan,-1}}$. 
% \amy{what does it mean to union two ranges? union is over sets}
\begin{equation}
    \label{interval_tan_ckw}
    \Theta_g^{\mt{tan,-1}} = [\phi^k-\pi/p+\pi/2, \quad \phi^k + \pi/p + \pi/2]
\end{equation}
\begin{equation}
    \label{interval_tan_antiCkw}
    \Theta_g^{\mt{tan,+1}} = [\phi^k - \pi/p + 3\pi/2, \quad \phi^k + \pi/p + 3\pi/2]
\end{equation}
Fig.~\ref{fig:angRangesTar} shows the different angular range sets for a target-robot interaction. 
It is worth mentioning that for noiseless sensors, if $z^{k-1}_g > z^{k+1}_g$ then $\Theta_g^{\mt{LOS}} = [\phi^k-\pi/p,\textrm{ }\phi^k]$. This results in a more accurate estimation of the location of a target and reduces the angular resolution error by half. The estimation of $\Theta_g^{\mt{tan}}$ and $\Theta_g^{\mt{avo}}$ also changes accordingly. 

As shown in our previous work \cite{selfPaper}, a heading direction in the angular ranges $\Theta_g^{\mt{LOS}}$ and $\Theta_g^{\mt{avo}}$ is guaranteed to make a robot move towards the target and away from the target, respectively. In contrast, a robot might end up moving towards or away from the target when it moves tangentially in an orbit.
Since secondary orbits are at least at a distance of $Or_0^{\mt{outer}}$ from a target, a robot moving tangentially in these orbits will always maintain a safe distance from the target. However, if a robot is moving tangentially in the primary orbit, we need to make sure that it maintains at least a distance of $Or_0^{\mt{inner}}$ from the target after moving $d_r$ units in the intended heading direction $\gamma_{r,T}$ such that $\theta_r \in \Theta_g^{\mt{tan}}$.

\begin{figure}[!ht]
  \centering
  \includegraphics[width=\linewidth]{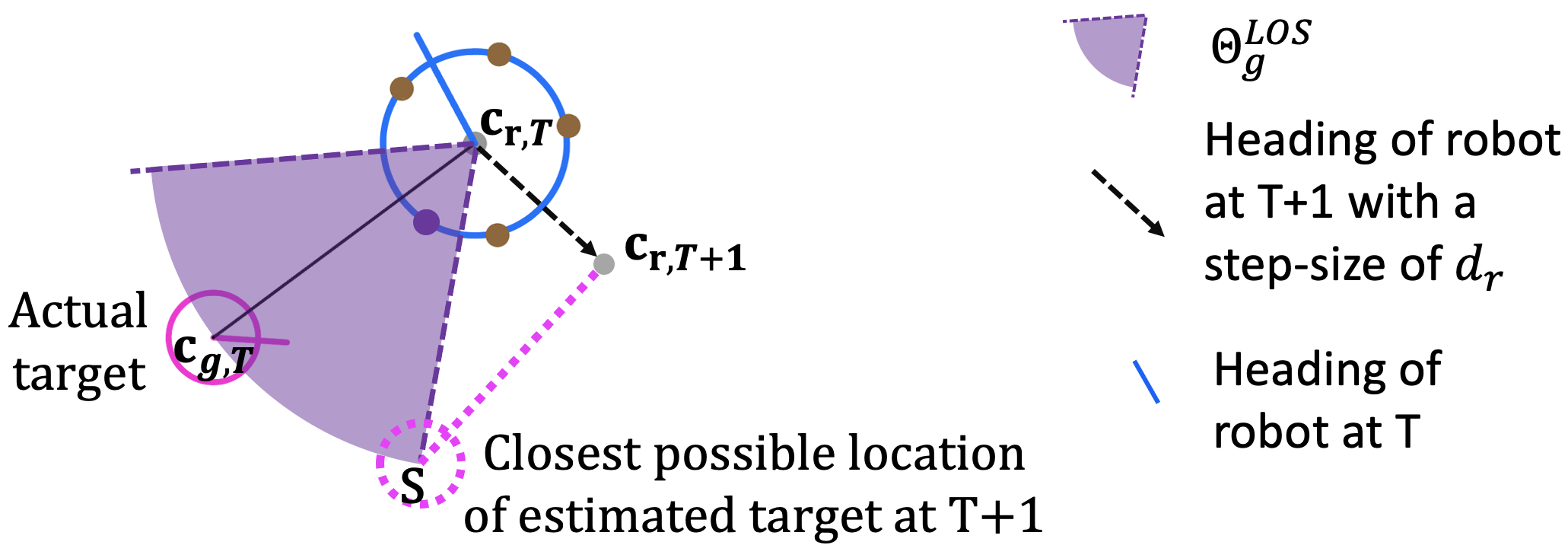}
    % \vspace{-1em}
  \caption{The distance $d_r$ that the robot can move in the intended heading is computed using the geometry of $\Delta\vb{S}\vb{c}_{r,T}\vb{c}_{r,T+1}$ %\hkc{might be worth laeling $d_r$ in the figure} 
%   \amy{shouldn't the black line be in the exact middle of $\Theta_g^{\mt{LOS}}$ }
% \himani{not necessarily, The black line is the actual LOS vector.It can be anywhere in the angular range cone}
  % \vspace{-1.3em}
  }
  \label{fig:tarLocate}
\end{figure}

In Fig.~\ref{fig:tarLocate}, we can see that at $T+1$, the closest possible location of the target is at $\vb{S} {\in \mathcal{I}}$.
% \hkc{what is $\mathcal{F}$? not the global frame? }
% \himani{$\mathcal{F}$ is the global frame}
% \hkc{i know. the notation is weird.}. 
If the heading direction $\theta_r \notin \Theta_g^{\mt{LOS}}$, the closest possible location of the target with respect to the robot's center at $T+1$ would be along one of the extremes of the angular range $\Theta_g^{\mt{LOS}}$.
To ensure safety, $\norm{\vb{c}_{r,T+1}-\vb{S}} \geq Or_0^{\mt{inner}}$. Using the {cosine rule of triangle} for $\triangle{\vb{S}\vb{c}_{r,T}\vb{c}_{r,T+1}}$, the {bounds on the} control parameter $d_r$ {can be computed using the} quadratic inequality given by Eq. ~\eqref{constraint_dist2tar} . 
%\hkc{I am not following the equation :/}
% where $\alpha =\min\limits_{\theta \in \Theta_g^{\mt{LOS}}}(\theta - \theta_r)$.
\begin{multline}
    \label{constraint_dist2tar}
    d_r^2 - 2d_r\norm{\vb{c}_{r,T}-\vb{S}}\mt{\cos}(\angle \vb{S}\vb{c}_{r,T}\vb{c}_{r,T+1}) \\ 
    + \norm{\vb{c}_{r,T}-\vb{S}}^2 {\geq (Or_0^{\mt{inner}})^2}
\end{multline}
{In the above inequality, $\norm{\vb{c}_{r,T}-\vb{S}}$ is under-approximated by the robot using the virtual source, as described in Section~\ref{section_vs}.}
% \hkc{again - too far from algo}In Algorithm \ref{algo:ca1}, the function \textbf{\texttt{DistAvoTar}} computes the value of $d_r$ from Eq. (\ref{constraint_dist2tar}).

\subsection{Local Control Law}
\label{section_controlSummary}
% \hkc{i would move the algo to be on the same page as Section IVD, the description}

\begin{algorithm}[ht]
\SetKwInOut{Input}{Input}
\SetKwInOut{Output}{Output}
\SetKwProg{Initialization}{Initialization}{}{}
\Input{$Z$, $B_s$, $p$, $d_r^{\mt{max}}$, $d_g^{\mt{max}}$, $r_s^{\mt{safe}}$, $\forall s \in \{r,g,e\}$, \texttt{orbits $Or_{i \geq 0}$}}
\Output{$d_r$, $\theta_r$}
 \tcp{compute $\Theta_e^\mt{avo}$, $\Theta_g^\mt{LOS}$, $\Theta_g^\mt{avo}$, $\Theta_g^\mt{tan}$, \texttt{currentOrbit}}
%  ,\texttt{currentOrbit}}
\If{\texttt{DistToEnvBound}$ \leq r_e^{\mt{safe}}+d_r^{\mt{max}}$}{
    $\theta_r$ = $\argmax\limits_{\theta \in \Theta_e^{\mt{avo}}}$ \texttt{DistAvoRob}$(Z_r,B_r,\theta,r_r^{\mt{safe}})$\\
    $d_r$ = \texttt{DistAvoRob}$(Z_r,B_r,\theta_r,r_r^{\mt{safe}})$\\
    % $\theta_r \in \Theta_e^{\mt{avo}}$ such that $d_r$ = $\max\limits_{\theta_r \in \Theta_e^{\mt{avo}}}$\texttt{DistAvoidRobots}$(Z_r,B_r,\theta_r,r_r^{\mt{safe}})$\\
}
\ElseIf{$\max(Z_g)=0$}{
% \tcp{do random walk while avoiding collision with other robots}
    $\theta_r$ = \texttt{randsample}($[0,\textrm{ } 2\pi]$)\\
    $d_r$ = \texttt{DistAvoRob}$(Z_r,B_r,\theta_r,r_r^{\mt{safe}})$\\
    \If{$d_r=0$}{
        $k= \texttt{argmin}(Z_r)$ \\
        $\theta_r$ = $\phi^k$\\
        $d_r$ = \texttt{DistAvoRob}$(Z_r,B_r,\theta_r,r_r^{\mt{safe}})$\\
    }  
}
\ElseIf{\texttt{DistToTar}$ < {Or_0^{\mt{inner}}}$}{
$\theta_r$ = $\argmax\limits_{\theta \in \Theta_g^{\mt{avo}}}$ \texttt{DistAvoRob}$(Z_r,B_r,\theta,r_r^{\mt{safe}})$\\
$d_r$ = \texttt{DistAvoRob}$(Z_r,B_r,\theta_r,r_r^{\mt{safe}})$\\
${d_r^{\mt{req}}}$ = $\norm{\textrm{\texttt{DistToTar}
% \hkc{should this be "Tar" and not "Target"?}
}-{Or_0^{\mt{inner}}}}$\\
\If{$d_r > d_r^{\mt{req}}$}{
$d_r = d_r^{\mt{req}}$\\
}
}
\ElseIf{\texttt{currentOrbit $ = Or_0$}}{
$\theta_r$ = $\argmax\limits_{\theta \in \Theta_g^{\mt{tan}}}$ 
$\min$\big(\texttt{DistAvoRob}$(Z_r,B_r,\theta,r_r^{\mt{safe}})$,
\hphantom{$\theta_r$ = arg max} \texttt{DistAvoTar} ($Z_g,B_g,\Theta_g^{\mt{LOS}},\theta)$\big)\\
% \hkc{it would be good to indent the following a bit. took me a while to parse}
$d_r$ = \texttt{DistAvoRob}$(Z_r,B_r,\theta_r,r_r^{\mt{safe}})$\\
}
\Else{
$\theta_r$ = $\argmax\limits_{\theta \in \Theta_g^{\mt{LOS}}}$ \texttt{DistAvoRob}$(Z_r,B_r,\theta,r_r^{\mt{safe}})$\\
$d_r$ = \texttt{DistAvoRob}$(Z_r,B_r,\theta_r,r_r^{\mt{safe}})$\\
\If{$d_r=0$}{
$\theta_r$ = $\argmax\limits_{\theta \in \Theta_g^{\mt{tan}}}$ \texttt{DistAvoRob}$(Z_r,B_r,\theta,r_r^{\mt{safe}})$\\
$d_r$ = \texttt{DistAvoRob}$(Z_r,B_r,\theta_r,r_r^{\mt{safe}})$\\
\If{$d_r=0$}{
    $k= \texttt{argmin}(Z_r)$ \\
    $\theta_r$ = $\phi^k$\\
    $d_r$ = \texttt{DistAvoRob}$(Z_r,B_r,\theta_r,r_r^{\mt{safe}})$\\
}  
}
}
\caption{Control algorithm for a robot}
\label{algo:ca1}
% \vspace{-0.4em}
\end{algorithm}

\textbf{Algorithm} \ref{algo:ca1} encodes the local reactive control law for a robot in the swarm that is tasked with searching and encapsulating targets while avoiding collisions. 

The algorithm describes the computation that happened at each time step $T$. Each robot in the swarm has: $Z$--the tuple of sensor measurements, $B_s$--the function describing the signal source strength as a function of radial distance from $s \in \{g,r,e\}$, the maximum step-size of a robot $d_r^{\mt{max}}$ and a target $d_g^{\mt{max}}$, the user-specified safety constraints for each source $r_s^{\mt{safe}}$, and the set \texttt{orbits} defined by an inner and outer radius of each orbit. 

The control synthesis proceeds as follows: First, the robot estimates its distance, \texttt{DistToEnvBound} (Section~\ref{section_vs}), from the environment boundary. If the robot is too close to the  boundary, it computes the allowed set of heading directions, $\Theta_e^{\mt{avo}}$. The direction of motion, $\theta_r$ is then chosen such that the robot moves away from the boundary with a maximum possible step size, $d_r$ while avoiding nearby robots (lines 1-2). {The function \textbf{\texttt{DistAvoRob}} computes this maximum possible value of $d_r$ from Eq.~\eqref{avoObsGivenTh}, as described in Section~\ref{section_ca}.} 
% \hkc{when you use the ref function, there should not be a space between the word and the $\Tilde{}$, otherwise it creates too big of a space. i fixed what I saw but please double check.}

Once the robot is at a safe distance from the boundary, it then estimates the relative distance \texttt{DistToTar} (Section~\ref{section_vs} ) from a target. 
If no target is sensed the robot performs a random walk while maintaining a safe distance from nearby robots (lines 4-10). 
If, on the other hand, the robot is inside the influence region of a target, it computes its current orbit, \texttt{currentOrbit} based on the estimated \texttt{DistToTar} and the input \texttt{orbits}. 
If the robot estimates that {the relative distance between itself and the target is}
% it is at a distance 
less than ${Or_0^{\mt{inner}}}$,
%\hkc{escape?or safe?}
% to the target
it computes the set of allowed heading direction, $\Theta_g^{\mt{avo}}$, and chooses a direction of motion from this set while maximizing the step size to avoid nearby robots (lines 11-16). The distance to move away from a target is capped at $d_r^{\mt{req}}=\norm{\textrm{\texttt{DistToTar}}-{Or_0^{\mt{inner}}}}$ (lines 15-16) to ensure that the robot doesn't move outside the primary orbit
%\hkc{how does this limit ensure that?}. 

If the robot is in the primary orbit (line 17), it moves tangentially to the orbit $Or_0$ (heading direction chosen from the computed set $\Theta_g^{\mt{tan}}$) with a step size $d_r$ such that it maintains a safe distance from the nearby robots and the target (lines 18-19). {The function \textbf{\texttt{DistAvoTar}} computes the maximum possible value of $d_r$ from Eq.~\eqref{constraint_dist2tar}, as described in Section~\ref{section_orbits}.} 

When a robot is in a secondary orbit, $Or_{i>0}$ it chooses a heading direction from the set $\Theta_g^{\mt{LOS}}$ to move towards the target while avoiding nearby robots (line 21-22). In case the robot cannot find a direction of motion to move a non-zero distance toward the target (line 23), it either moves a non-zero distance tangentially in its current orbit (lines 24-25) or moves a safe distance in a heading direction based on the reading from the sensor receiving the minimum signal strength $z_r^k$, i.e. the direction where the virtual source corresponding to other robots is the farthest (lines 26-29). 

A robot's local control, as summarized in Algorithm \ref{algo:ca1}, is agnostic to the motion type of the targets. This, together with our convergence guarantees in the following section, implies that our algorithm guarantees the encapsulation of multiple targets moving in the bounded environment with different types of motion models, as described in Section~\ref{section_targetMotion}. 

% The function \texttt{DistToAvoTarget} in 
% \textbf{Algorithm} \ref{algo:ca1} uses the above equation to compute the distance that a robot can safely move with respect to target. In \textbf{Algorithm} \ref{algo:ca1}, the functions \texttt{DistToEnvBound} and \texttt{DistToTarget} are used to compute these relative distances for the environment boundary and a target respectively. In Algorithm [1], the function \texttt{DistAvoidRobots} is used to compute the maximum possible value of $d_r$ from Eq. ~\eqref{avoObsGivenTh}.

% \subsection{Local control law for a robot}
% - summarize the "if-else" control law \\
% - include algorithm block for the control of a robot
% \label{section_controlSummary}
%%%%%%%%%%%%%%%%%%%%%%%%%%%%%%%%%%%%%%%%%%%%%%%%%%%%%%%%%%%%%%%%%%%%%%%%
\section{Convergence Guarantees}
\label{sec_proof}
We use the Lyapunov stability theory to provide guarantees on the emergent behavior of the swarm. In this section, for clarity, we consider circular robots
% operating synchronously (i.e. the same clock) and 
with noiseless sensors. 
{In practice, the desired behavior emerges}
% Though, the convergence guarantees hold 
for non-circular robots as well, which we demonstrate in simulations in Section~\ref{sec_sim}.
% \hkc{not sure I would say that because you are not proving. maybe instead of "Though, the convergence guarantees hold" write "In practice, we see the desired behavior emerge..."}

\label{guar_live}
\begin{lemma}
\label{lemma_RWexplore}
From~\cite{popov_2021}: A disc robot with a non-zero radius performing a random walk in a bounded 2D environment will always eventually explore the entire area.
\end{lemma}
{\begin{lemma}
    \label{lemma_colavo_tar}
    For any arbitrary initial condition such that a robot is at least $r_g^{\mt{safe}}$ away from a target, a necessary condition to ensure a collision-free target's motion is that the escape radius of the target,  $r_g^{\mt{escape}} \geq r_g^{\mt{safe}} + d_g^{\mt{max}}$ and the inner radius of the primary orbit, $Or_0^{\mt{inner}} \geq r_g^{\mt{safe}} + d_g^{\mt{max}}$
\end{lemma}
\begin{proof}
The lower bound on $r_g^{\mt{escape}}$ ensures that a target gets enough margin to escape an approaching robot.
As described in Section~\ref{section_orbits}, a robot's behavior is such that it moves away from the target if the robot crosses the inner ring, $Or_0^{\mt{inner}}$, of the primary orbit. Hence the above lower bound on $Or_0^{\mt{inner}}$ ensures that collision avoidance behavior for a robot is triggered before the distance  between a target and a robot becomes $r_g^{\mt{safe}}$. 
% \hkc{why is (ii) dependent on the escape region?}.    
\end{proof}
}

\begin{lemma}
\label{lemma_rencap}
For any arbitrary initial condition such that a robot is at least $r_g^{\mt{safe}}$ away from a target the following are the necessary
% \hkc{necessary or sufficient?}
conditions to ensure a target's encapsulation:
% \hkc{it is conditions on "collision free", not on encapsulation. maybe change "encapsulation" to "motion"?},
% \himani{These are the necessary conditions for encapsulation. Sufficiency conditions also depend on lemma V.3-V.5 }
\begin{enumerate}
    \item the outer radius of the encapsulation ring $\mathcal{A}_g$ satisfies  
    \begin{multline}
        r_g^{\mt{encap}} \geq d_r^{\mt{max}} + r_r \\+  \sqrt{(Or_0^{\mt{inner}})^2 + r_r^2 - 2r_rOr_0^{\mt{inner}}\mt{cos}(\pi/p)} 
    \end{multline}
    \item the number of robots $n_g$ specified for encapsulation satisfies, 
    \begin{equation}
        \label{totalRobs}
        n_g \leq n_0 = \frac{2\pi}{\mt{cos}^{-1}\bigg(1 - \frac{(\beta_r + r_r)^2}{2(r^\mt{encap}_g)^2}\bigg)} 
    \end{equation}
\end{enumerate}
\end{lemma}
\begin{proof}
Each robot's estimate of the relative distance from a target depends on the sensor with the maximum reading, $\max(Z_g)$. 
% \hkc{why have this sentence: ?}If we consider a robotic swarm with asymmetric sensor placement, then this estimate can differ for each robot. 
Given that a virtual source is always either closer or at the radial location of an actual source, it is possible that even if a robot is present in the primary orbit,
% in the inertial frame $\mathcal{I}$,
{the robot estimates itself to be present at a relative distance of less than $Or_0^{\mt{inner}}$ with respect to the target.} 
% \hkc{why? $->$}
This will trigger the collision avoidance behavior for the robot and it will move away from the target.
% \hkc{there is something implicit about the escape domain that needs to be clarified. I have a bunch of comments about in prior sections}
To successfully encapsulate a target $g$, it is required that the outer radius of the encapsulation ring, $r_g^{\mt{encap}}$ incorporate the robot with the worst possible estimate of the target's location. This will ensure that the robot remains in the primary orbit even after being over-cautious in moving away from the target. 
% This is because a robot starts to encircle a target only when it estimates its \textit{current orbit} to be the primary orbit.

At each time step, a robot chooses its control parameters such that it maintains at least a distance of $Or_0^{\mt{inner}}$ from a target, that is, $\norm{\vb{c}_{g,T}-\vb{c}_{r,T}} \geq Or_0^{\mt{inner}}$. 
Since $Or_0^{\mt{inner}}$ is defined between a robot's center and the target, we set $d_s = Or_0^{\mt{inner}}$ in Eq. ~\eqref{dsk} to obtain $d_g^k = \sqrt{(Or_0^{\mt{inner}})^2 + r_r^2 - 2r_rOr_0^{\mt{inner}}\mt{cos}(\pi/p)}$. A robot will start to move away from the target when $\mt{max}(Z_g) \geq B_g(d_g^k)$. 
At this point, the upper bound on $\norm{\vb{c}_{g,T}-\vb{c}_{r,T}}$ is given by Eq. ~\eqref{orbitInnerDist}. 
We can see that as $p \rightarrow \infty$, $\norm{\vb{c}_{g,T}-\vb{c}_{r,T}} \rightarrow Or_0^{\mt{inner}}$ and for a finite $p$, collision avoidance behavior is triggered before the robot is at a distance of $Or_0^{\mt{inner}}$ from the target. 
\begin{equation}
\label{orbitInnerDist}
    \norm{\vb{c}_{g,T}-\vb{c}_{r,T}} \leq r_r +  \sqrt{(Or_0^{\mt{inner}})^2 + r_r^2 - 2r_rOr_0^{\mt{inner}}\mt{cos}(\pi/p)}
\end{equation} 
For asymmetric sensor placement, we replace $\pi/p$ with half of the maximum angle between two adjacent sensors on the robot.

To incorporate the robot with the worst estimate of a target's location, we set a lower bound on the outer radius of the encapsulation ring using Eq. ~\eqref{orbitInnerDist} as $r_g^{\mt{encap}} \geq r_r +  \sqrt{(Or_0^{\mt{inner}})^2 + r_r^2 - 2r_rOr_0^{\mt{inner}}\mt{cos}(\pi/p)}$.
Since the robot may chatter in the encapsulation ring due to constant attraction and repulsion from the target and nearby robots, we add $d_r^{\mt{max}}$ to the lower bound on $r_g^{\mt{encap}}$ (condition 1). This will ensure that a robot remains in the encapsulation ring when there are robots nearby. 
% We set $r_g^{\mt{encap}}$ as the outer radius $Or_0^{\mt{outer}}$ of the primary orbit for a robot. 
% At each time-step, the control parameters of a robot are chosen such that it moves towards the primary orbit $Or_0^{\mt{inner}}$ and thereafter moves tangentially in it. If a robot crosses $Or_0$ it moves away from the target. That is, it always maintains at least $r_g^{\mt{safe}}$ distance from a target.

Furthermore, the maximum number of robots that can be specified for target encapsulation  (condition 2) is bounded by the total number of robots that can be physically placed in the encapsulation ring such that the encapsulating robots are outside each other's influence region to ensure no chattering that can be caused by repulsion from each other. 
% This is shown in Fig. \ref{fig:boundN0} and computed in condition (2).
% \amy{for a proof, do you need to actually derive the equation? maybe your explanation is sufficient, i'm not sure}
% \begin{figure}
%   \centering
%   \includegraphics[width=0.4\linewidth]{figures/boundN0.png}
%   \caption{Total robots $n_g$ in the primary orbit should be such that they are outside each other's influence region.}
%   \label{fig:boundN0}
%   \Description{Total robots in the primary orbit should be such that they are outside each other's influence region.}
% \end{figure}
When $n_g > n_0$ a dynamic equilibrium exists around a target such that there are always almost $n_0$ robots present in the primary orbit \cite{selfPaper}.
\end{proof}

As described in Section~\ref{section_problemForm}, we consider three types of motion patterns that a target can exhibit.
For each of these target motion patterns, we provide guarantees for liveness (eventual encapsulation) based on the Lyapunov stability theory and stochastic analysis (Lemmas~\ref{lemma_randTarEscape} -~\ref{lemma_predeterTarEscape}). 

\begin{lemma}
\label{lemma_randTarEscape}
Consider a target $g \in \mathcal{G}$ moving randomly in the bounded environment until it senses any robot in its escape domain (as described by motion model~\ref{randMotionEscape} in Section~\ref{section_targetMotion}). If,
\begin{enumerate}
    \item the maximum step size of the target, $d_g^{\mt{max}} \leq \lambda d_r^{\mt{max}}$ where 
    \begin{align*}
        \lambda &= \mt{min}\bigg(\frac{\pi}{2},\textrm{  }\frac{\alpha}{\mt{sin}(\pi-\alpha)}\bigg)\frac{\mt{sin}\varphi}{\varphi}\mt{cos}\varphi \\
        \alpha &= \mt{cos}^{-1}\bigg(1 - \frac{(\beta_r + r_r)^2}{2(r_g^{\mt{escape}})^2}\bigg)\\
        \varphi &= \frac{\max(\phi^k - \phi^{k+1})}{2},\quad k = \{1 \cdots p\}\\
        &= \pi/p, \quad \mt{for \textrm{ }  symmetric \textrm{ } sensor \textrm{ }  placement}
    \end{align*}
    % \amy{what is $\varphi$ for asym sensor placement? is it just max(sensor placement dist)?}
\end{enumerate}
the target $g$ will be encapsulated eventually.
\end{lemma}
\begin{proof}
Consider a target $g \in \mathcal{G}$.
Let $\vb{u}_g = [d_g\mt{cos}(\gamma_{g,T} + \theta_g)\quad d_g\mt{sin}(\gamma_{g,T} + \theta_g)]$ and $\vb{u}_r = [d_r\mt{cos}(\gamma_{r,T} + \theta_r)\quad d_r\mt{sin}(\gamma_{r,T} + \theta_r)]$
% \hkc{the notation here does not match the figure} 
be the control input of a target and robot respectively at time $T$, and $\eta$ be the total robots currently present in the \textit{escape} domain of the target, that satisfy $\norm{\vb{c}_{g,T}-\vb{c}_{r,T}} \leq r_g^{\mt{escape}}$.
As outlined in Section~\ref{section_orbits}, the motion strategy of a robot can be broken down as follows:
\newline
\noindent\textbf{Case I: Robot is in an orbit $Or_{i \geq 0}$ such that $\eta = 0$}\\
We use the definition of stochastic stability in the sense of Lyapunov \cite{stochasticDiscrete_li} to show that a robot eventually reaches the primary orbit $Or_0$.
Let $V = \norm{\vb{c}_{g,T}-\vb{c}_{r,T}}^2$ be the candidate Lyapunov function defined on the the domain $D_r \subseteq \mathbb{R}^2$ such that $\norm{\vb{c}_{g,T}-\vb{c}_{r,T}} \geq r_g^{\mt{encap}}$. 
Using Eq. ~\eqref{targetModel} and Eq. ~\eqref{robotModel} we have, $\Delta V = \norm{(\vb{c}_{g,T} + \vb{u}_g) - (\vb{c}_{r,T}+\vb{u}_r)}^2 - \norm{\vb{c}_{g,T}-\vb{c}_{r,T}}^2$. 
For ease of exposition, we will drop the subscript $T$ in the following analysis. 
On simplifying, ${\Delta V} = \norm{\vb{u}_g-\vb{u}_r}^2 + 2(\vb{c}_g-\vb{c}_r) \cdot (\vb{u}_g-\vb{u}_r) $.
\begin{figure}[!ht]
  \centering  \includegraphics[width=0.8\linewidth]{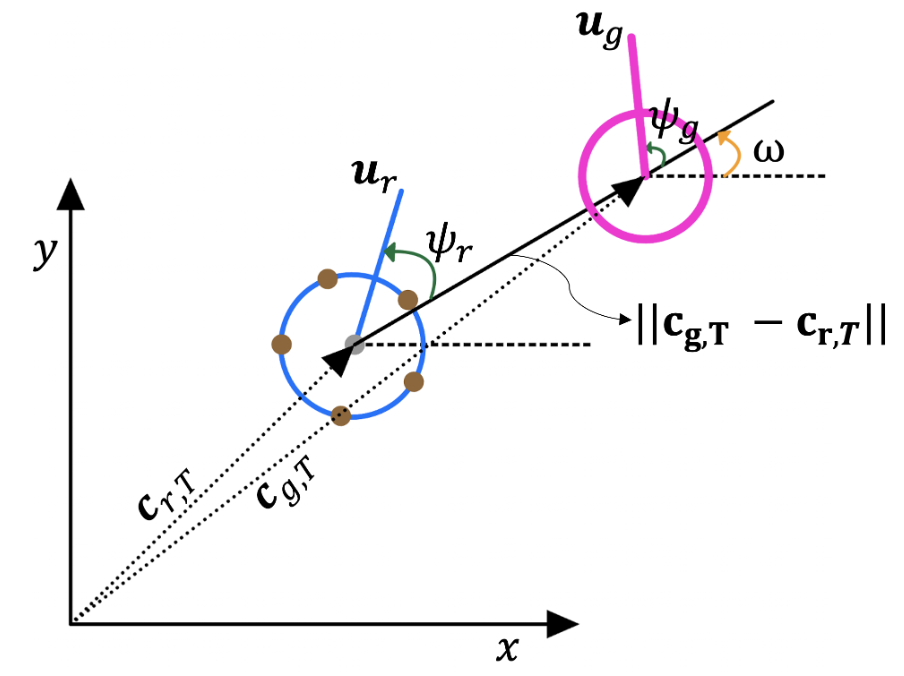}
  \caption{Relative kinematics of a robot-target interaction.
  % \hkc{what is $\psi$? it does not show up in the proof [ok, it does, but way later]}
  }
  \label{fig:relativeTarRob}
\end{figure}
{Fig.~\ref{fig:relativeTarRob} depicts the relative kinematics model between the target and a robot where $\omega$ is the angle that the LOS vector, ($\vb{c}_g-\vb{c}_r$) makes with x-axis.}
Let $\hat{\vb{l}} = [\mt{cos}\omega\quad\mt{sin}\omega]$
% \hkc{why are you using 'T'? does it matter if it is a row or column? it is confusing with the timestep that is also 'T'} 
be the vector along $(\vb{c}_g-\vb{c}_r)$ and $\hat{\vb{t}} = [-\mt{sin}\omega\quad\mt{cos}\omega]$ be the vector tangential to it.
% where $\omega$ is the angle that the LOS vector, ($\vb{c}_g-\vb{c}_r$) makes with x-axis as shown in Fig. \ref{fig:relativeTarRob}. 
Then,
\begin{align}
\label{deltaV}
    {\Delta V} &= d_g^2+d_r^2-2\vb{u}_g\cdot\vb{u}_r + 2\norm{\vb{c}_g-\vb{c}_r}\hat{\vb{l}} \cdot (\vb{u}_g-\vb{u}_r)
\end{align}

\noindent\textbf{(a) Robot is in a secondary orbit:}
For this case, a robot would move towards the target, that is $\theta_r \in \Theta_g^{\mt{LOS}}$ given in Eq. ~\eqref{interval_of_tar}.
If $\eta = 0$, that is, there are no robots in the target's escape domain, the target moves randomly. Hence, $\theta_g \in [0 \textrm{  } 2\pi)$.
That is, both $\theta_g$ and $\theta_r$ are stochastic. 
Moreover, the control inputs $\vb{u}_g$ and $\vb{u}_r$ are independent random vectors and their corresponding expected values are given by, 
\begin{align}
    \nonumber
    \mathbb{E}[\vb{u}_g]& = d_g\mathbb{E}[\mt{cos}(\gamma_g+\theta_g) \textrm{ } \mt{sin}(\gamma_g+\theta_g)] \\ 
    \nonumber
    &= d_g\Bigg[\int\limits_{0}^{2\pi}\mt{cos}(\gamma_g + \theta_g)\frac{1}{2\pi}\mt{d}\theta_g \quad \int\limits_{0}^{2\pi}\mt{sin}(\gamma_g + \theta_g)\frac{1}{2\pi}\mt{d}\theta_g\Bigg] \\
    \label{avg_ug}
    &= 0 
    \\ \nonumber
% \end{align}
% \begin{align} 
    \nonumber
    \mathbb{E}[\vb{u}_r] &= d_r\mathbb{E}[\mt{cos}(\gamma_r+\theta_r) \quad \mt{sin}(\gamma_r+\theta_r)]
    \\
% \end{align}
% \begin{align}
    \nonumber
    &= d_r\Bigg[\int\limits_{\phi_k-\frac{\pi}{p}}^{\phi_k+\frac{\pi}{p}}\mt{cos}(\gamma_r + \theta_r)\frac{1}{2\pi/p}\mt{d}\theta_r 
    % \quad
    \\ \nonumber
    % \end{align}
    % \begin{align}
    \nonumber
    &\quad \quad \quad \quad \quad \quad \quad 
    \int\limits_{\phi_k-\frac{\pi}{p}}^{\phi_k+\frac{\pi}{p}}\mt{sin}(\gamma_r + \theta_r)\frac{1}{2\pi/p}\mt{d}\theta_r\Bigg] 
    % \\ \nonumber
\end{align}
\begin{align}
    \nonumber
    &= d_r\frac{\mt{sin}\varphi}{\varphi}[\mt{cos}(\gamma_r+\phi^k) \quad \mt{sin}(\gamma_r+\phi_k)]\\
    \label{avg_ur} 
    &= d_r\frac{\mt{sin}\varphi}{\varphi}\hat{\vb{u}}^k_r
\end{align}
% \begin{align}
% \label{avg_ur}
%     \nonumber
%     &= d_r\frac{\mt{sin}\varphi}{\varphi}[\mt{cos}(\gamma_r+\phi^k) \quad \mt{sin}(\gamma_r+\phi_k)]\\
%     &= d_r\frac{\mt{sin}\varphi}{\varphi}\hat{\vb{u}}^k_r
% \end{align}
where $\varphi = \pi/p$ for symmetric sensor placement and $\hat{\vb{u}}^k_r$ is the unit vector in the direction of $k^{\mt{th}}$ sensor. 
Intuitively this means that on an average the robot moves in the direction of the $k^{\mt{th}}$ sensor (receiving maximum intensity from the target) with a step-size reduced by the factor $\mt{sin}\varphi/\varphi$. Furthermore, as $p \rightarrow \infty$, $\mathbb{E}[\vb{u}_r] \rightarrow d_r\hat{\vb{u}}_r^k$. That is, if the robot knows the \textit{exact} relative location of the target, it moves towards the target along the line of sight vector with the maximum possible step size.

Using Eq. ~\eqref{avg_ug} and Eq. ~\eqref{avg_ur}, the expected value of change in the Lyapunov function (as given by Eq. ~\eqref{deltaV}) between two consecutive time steps is,
\begin{align}
    \nonumber
    \mathbb{E}[\Delta V] &= d_g^2+d_r^2+ 2\mathbb{E}\big[\norm{\vb{c}_g-\vb{c}_r} \hat{\vb{l}}\cdot (\vb{u}_g-\vb{u}_r)\big] \\ \nonumber
    &=d_g^2+d_r^2+ 2\norm{\vb{c}_g-\vb{c}_r}\big(\mathbb{E}[ \vb{u}_g]- \mathbb{E}[\vb{u}_r]\big)\cdot \hat{\vb{l}} \\
% \end{align}
% Substituting Eq. ~\eqref{avg_ur},
% \begin{align}
    \label{lyaProof_escape_contd}
    % \nonumber
    %  \mathbb{E}[\Delta V] &= 
    %  d_g^2+d_r^2-2\mathbb{E}[\vb{u}_g]\cdot\mathbb{E}[\vb{u}_r] + 2\norm{\vb{c}_g-\vb{c}_r}\mathbb{E}[\hat{\vb{l}} \cdot (\vb{u}_g-\vb{u}_r)] \\
     &= d_g^2+d_r^2- 2\norm{\vb{c}_g-\vb{c}_r} d_r\frac{\mt{sin}\varphi}{\varphi}\hat{\vb{u}}^k_r \cdot \hat{\vb{l}}  
\end{align}
% For stability, we require that $\mathbb{E}[\Delta V] \leq 0$. 
% The necessary condition to ensure stability is $\hat{\vb{u}}^k_r \cdot \hat{\vb{l}} \geq 0$. 
The maximum deviation of the unit vector in the direction of the $k^{\mt{th}}$ sensor, $\hat{\vb{u}}^k_r$ from the LOS vector $\hat{\vb{l}}$ is limited to $\varphi = \pi/p$ (refer to Fig.~\ref{fig:vs_iros}), that is $\hat{\vb{u}}^k_r \cdot \hat{\vb{l}} \geq \mt{cos}\varphi$. Furthermore, when a robot is in a secondary orbit $\norm{\vb{c}_g-\vb{c}_r} \geq r_g^{\mt{encap}}$. Substituting these bounds in Eq. ~\eqref{lyaProof_escape_contd} we have,
\begin{align*}
    \mathbb{E}[\Delta V] \leq  d_g^2+d_r^2- 2r_g^{\mt{encap}} d_r\frac{\mt{sin}\varphi}{\varphi} \mt{cos}\varphi
\end{align*}
For stability, we require that $\mathbb{E}[\Delta V] \leq 0$. That is
\begin{align}
\label{lyaProof_final}
    d_r^2- \big(2r_g^{\mt{encap}}\frac{\mt{sin}\varphi}{\varphi}\mt{cos}\varphi\big) d_r + d_g^2 \leq 0
\end{align}
The necessary condition to satisfy the above inequality is that $\mt{cos}\varphi \geq 0$. That is, the total number of sensors on a robot must be greater than or equal to three.
The roots of the above quadratic inequality in $d_r$ give us an upper bound on the maximum step-size of a robot $d_r^{\mt{max}}$ which is always larger than the bound determined in Eq.~\eqref{bound_d_r_max} and hence Eq. ~\eqref{lyaProof_final} is always satisfied. 
\\

\noindent\textbf{(b) Robot is in the primary orbit:}
Once a robot reaches the primary orbit, it moves tangentially to the orbit while ensuring that $\norm{\vb{c}_{g}-\vb{c}_{r}} \geq Or_0^{\mt{inner}}$.
% \hkc{still not sure about why we care about the escape region here}
% A robot moves tangentially in a orbit if it cannot move towards the target due to the presence of other robots in LOS or if it is in primary orbit such that $\norm{\vb{c}_g-\vb{c}_r}> r_g^{\mt{escape}}$.
To analyze this, we look at how the LOS vector between a target and robot changes between two time steps, which is given by $\Delta (\vb{c}_g-\vb{c}_r) = \vb{u}_g - \vb{u}_r$.
% This is given by Eq. ~\eqref{deltaLOS}.
% \begin{align}
%     \label{deltaLOS}
%     \Delta (\vb{c}_g-\vb{c}_r) = \vb{u}_g - \vb{u}_r
% \end{align}
In Eq.~\eqref{relVel_gr_rad} and Eq.~\eqref{relVel_gr_tan} we define $u^{\mt{rad}}_{gr}$ and $u^{\mt{tan}}_{gr}$ representing the polar coordinates corresponding to the radial and tangent component of the change in LOS vector in global frame $\mathcal{I}$. 
\begin{align}
    \label{relVel_gr_rad}
    u^{\mt{rad}}_{gr} &=  (\vb{u}_g - \vb{u}_r) \cdot \hat{\vb{l}} \\ 
    \label{relVel_gr_tan}
    u^{\mt{tan}}_{gr} &= (\vb{u}_g - \vb{u}_r) \cdot \hat{\vb{t}}
\end{align}
As explained earlier, to ensure a target's encirclement, it is necessary that a robot is able to complete a revolution around the target in the primary orbit $Or_0$.
% \amy{why?}. 
We evaluate this stochastically by computing the expected value of the change in the tangential component of the relative LOS vector, $    \mathbb{E}[u_{gr}^{\mt{tan}}] = \mathbb{E}[\vb{u}_g]\cdot \hat{\vb{t}} - \mathbb{E}[\vb{u}_r]\cdot \hat{\vb{t}}$.
% \begin{align*}
%     \mathbb{E}[u_{gr}^{\mt{tan}}] = \mathbb{E}[\vb{u}_g]\cdot \hat{\vb{t}} - \mathbb{E}[\vb{u}_r]\cdot \hat{\vb{t}}
% \end{align*}
For a clockwise orbital rotation, $\theta_r \in \Theta_g^{\mt{tan},-1}$. Since $\eta = 0$ for this case, $\theta_g \in [0,2\pi)$. Simplifying and substituting Eq. ~\eqref{avg_ug} in the above equation we have
\begin{multline}
\nonumber
    \mathbb{E}[u_{gr}^{\mt{tan}}] =  -d_r\Bigg[\int\limits_{\phi_k-\frac{\pi}{p}+\frac{\pi}{2}}^{\phi_k+\frac{\pi}{p}+\frac{\pi}{2}}\mt{cos}(\gamma_r + \theta_r)\frac{1}{2\pi/p}\mt{d}\theta_r \quad \\ \int\limits_{\phi_k-\frac{\pi}{p}+\frac{\pi}{2}}^{\phi_k+\frac{\pi}{p}+\frac{\pi}{2}}\mt{sin}(\gamma_r + \theta_r)\frac{1}{2\pi/p}\mt{d}\theta_r\Bigg] \cdot \hat{\vb{t}}
\end{multline}
\begin{align}
\label{robTan_avg}
    \nonumber
    &= -d_r\frac{\mt{sin}\varphi}{\varphi}[-\mt{sin}(\gamma_r+\phi^k) \quad \mt{cos}(\gamma_r+\phi_k)]\cdot \hat{\vb{t}} \\
    &= -d_r\frac{\mt{sin}\varphi}{\varphi} \hat{\vb{u}}^k_r \cdot \hat{\vb{l}}
    % &= -d_r\frac{\mt{sin}\varphi}{\varphi}\mt{cos}(\varphi)
\end{align}
Eq. ~\eqref{robTan_avg} shows that a robot moves clockwise in the primary orbit with an expected tangential step-size of $d_r\frac{\mt{sin}\varphi}{\varphi}\mt{cos}(\varphi)$ with respect to the target.
% From Eq. ~\eqref{robTan_avg}, we can see that, on average, a robot moves clockwise in the primary orbit with at least a step-size of $d_r\frac{\mt{sin}\varphi}{\varphi}\mt{cos}(\varphi)$.
% \amy{what does it mean to be "on average with at least a step-size of"}
\\

\noindent\textbf{Case II: $\norm{\vb{c}_{g,T}-\vb{c}_{r,T}} \leq r_g^{\mt{escape}}$}\\
Since a robot orbiting in $Or_0$ also avoids nearby robots, at a timestep $T$, it is possible that a robot is marginally outside the  escape domain of the target but is unable to move in $\Theta_g^{\mt{avo}}$ due to the presence of other robots in orbit $Or_1$ at a distance of $r_r^{\mt{safe}}$. 

This behavior could lead to
$\norm{\vb{c}_{g,T+1}-\vb{c}_{r,T+1}} < r_g^{\mt{escape}}$, resulting in $\eta > 0$. The target would then move such that it can escape from all the robots present in its escape domain. Let $\psi_g$ be the angle between a target's intended heading and the LOS vector ($\vb{c}_{g,T}-\vb{c}_{r,T}$), as shown in Fig.~\ref{fig:relativeTarRob}. If $\eta = 1$, then $\psi_g \in [3\pi/2,\textrm{  } \pi/2]$. % \amy{why do we have a new variable $\psi$?}. 
If $\eta = 2$ {and $\alpha$ is the angle that these robots subtend at the center of the target, as shown in Fig.~\ref{fig:tarEscape},} then $\psi_g \in [3\pi/2 + \alpha,\textrm{  } \pi/2]$ or $\psi_g \in [3\pi/2,\textrm{  } \pi/2 - \alpha]$, depending on which robot $\psi_g$ is measured with respect to. 
% \hkc{i don't understand "is measured with" - what is it referring to? also, what is $\alpha$?[oh, i see it is described next. might be worth reversing the order in the writing here]}. 
Without loss of generality, we can consider one of these. 
{That is, }if $\eta > 1$, the available angular range of the target for escaping decreases from $\pi$ to ($\pi - (\eta - 1)\alpha$). 

To determine $\alpha$, we use the fact that robots in the primary orbit disperse such that, on average, they are outside each other's influence region. Then, using geometry shown in Fig.~\ref{fig:tarEscape}, $\alpha = \mt{cos}^{-1}\bigg(1 - \frac{(\beta_r + r_r)^2}{2(r_g^{\mt{escape}})^2}\bigg)$. 
Note that, the maximum escaping angular range of a target is limited to $\pi$ (when $\eta = 1$). Hence, for $\eta > \pi/\alpha$, the target can no longer escape with the maximum possible step size. 
\begin{figure}[!ht]
  \centering
  \includegraphics[width=0.75\linewidth]{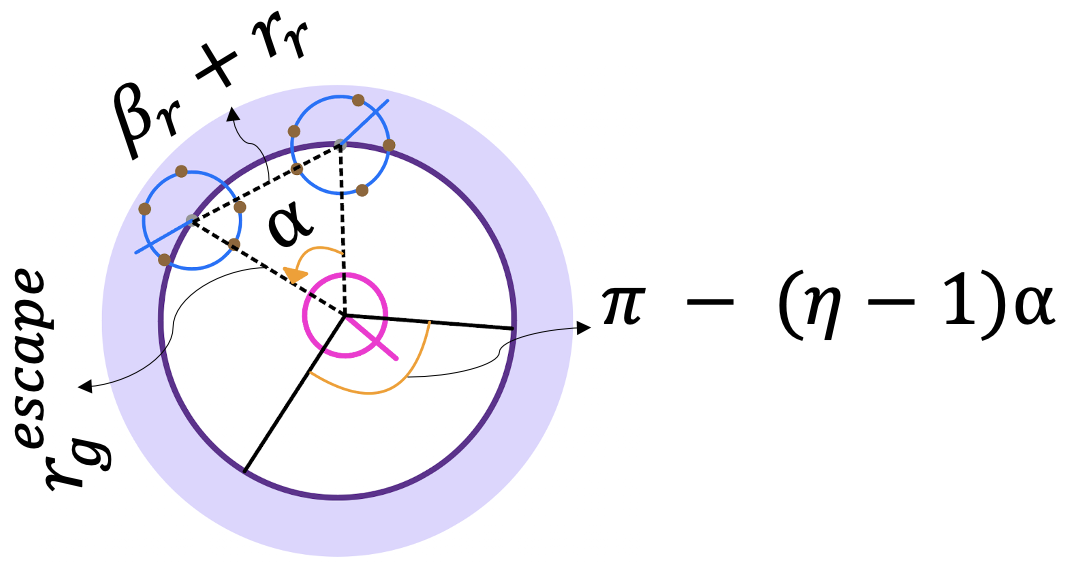}
    % \vspace{-1.3em}
  \caption{Geometric configuration for computing bounds on the ratio between target and robot step-sizes.}
  \label{fig:tarEscape}
  % \vspace{-1.3em}
\end{figure}
For \textbf{Case II}, we have to ensure that (i) robots implementing tangential control law in the primary orbit are able to encircle the target, that is $\mathbb{E}[u_{gr}^{\mt{tan}}] \leq 0$ if $\theta_r \in \Theta_g^{\mt{tan},-1}$ and $\mathbb{E}[u_{gr}^{\mt{tan}}] \geq 0$ if $\theta_r \in \Theta_g^{\mt{tan},1}$, and 
(ii) robots in secondary orbits are able to move towards the target, that is $\mathbb{E}[u_{gr}^{\mt{rad}}] \leq 0$ for $\theta_r \in \Theta_g^{\mt{LOS}}$. 

Using Eq. ~\eqref{relVel_gr_tan} and the tangential control input ($\theta_r \in \Theta_g^{\mt{tan},-1}$) for robots present in $Or_0$ we have, 
% \amy{should the first line of the equation have the dot product operator?}
\begin{align*}
    \mathbb{E}[u_{gr}^{\mt{tan}}] &= d_g\mathbb{E}[\mt{sin}\psi_g]-\mathbb{E}[\vb{u}_r] \cdot \hat{\vb{t}} \\
    &= d_g\int\limits_{\frac{3\pi}{2} + (\eta-1)\alpha}^{\frac{\pi}{2}}\mt{sin}\psi_g\frac{1}{\pi-(\eta-1)\alpha} \, \mt{d}\psi_g - \mathbb{E}[\vb{u}_r]\cdot\hat{\vb{t}} \\
    &= d_g\frac{\mt{sin}((\eta-1)\alpha)}{(\pi-(\eta-1)\alpha)}-d_r\frac{\mt{sin}\varphi}{\varphi}\hat{\vb{u}}_r^k\cdot\hat{\vb{t}}
\end{align*}
To ensure clockwise orbital rotation, $\mathbb{E}[u_{gr}^{\mt{tan}}] \leq 0$, that is, 
\begin{align}
\label{dgmax_bound_tan}
\nonumber
    \max\limits_{\eta \leq \pi/\alpha}\Bigg(d_g\frac{\mt{sin}((\eta-1)\alpha)}{(\pi-(\eta-1)\alpha)}\Bigg) &\leq \min\limits_{\hat{\vb{u}}_r^k\cdot \hat{\vb{t}} \geq \mt{cos}\varphi}\Bigg(d_r\frac{\mt{sin}\varphi}{\varphi}\hat{\vb{u}}_r^k\cdot \hat{\vb{t}}\Bigg)\\
    d_g &\leq \frac{\alpha}{\mt{sin}(\pi-\alpha)}d_r\frac{\mt{sin}\varphi}{\varphi}\mt{cos}\varphi
\end{align}
% \amy{from my understanding, second eqn is a simplification of the first. if so, you should mention that (right now it looks like they are two completely independent eqns)}
Intuitively, $\mathbb{E}[\vb{u}_g\cdot \hat{\vb{t}}]$ is maximal when the target has the least freedom in choosing its motion, that is $\eta = \pi/\alpha$. Similarly, $\mathbb{E}[\vb{u}_r\cdot \hat{\vb{t}}]$ is minimal when the average heading direction $\hat{\vb{u}}_r^k$ is deviated the most from $\hat{\vb{t}}$.
% \amy{wording is weird. do you mean to say the control is perpendicular to the tangential direction?}. 

Now, we need to ensure that the robots in the secondary orbits move toward an escaping target. Using Eq. ~\eqref{relVel_gr_rad} and the LOS control input ($\theta_r \in \Theta_g^{\mt{LOS}}$) for robots present in $Or_{i>0}$ we have,
% \amy{dots here again}
\begin{align*}
    \mathbb{E}[u_{gr}^{\mt{rad}}] &= \mathbb{E}[\vb{u}_g \cdot \hat{\vb{l}}]-\mathbb{E}[\vb{u}_r] \cdot \hat{\vb{l}} \\
    &= d_g\int\limits_{\frac{3\pi}{2} + (\eta-1)\alpha}^{\frac{\pi}{2}}\mt{cos}\psi_g\frac{1}{\pi-(\eta-1)\alpha} \, \mt{d}\psi_g - \mathbb{E}[\vb{u}_r]\cdot\hat{\vb{l}} \\
    &= d_g\frac{1+\mt{cos}((\eta-1)\alpha)}{(\pi-(\eta-1)\alpha)}-d_r\frac{\mt{sin}\varphi}{\varphi}\hat{\vb{u}}_r^k\cdot \hat{\vb{l}}
\end{align*}
The distance between a target and robot will decrease if $\mathbb{E}[u_{gr}^{\mt{rad}}] \leq 0$. That is,
\begin{align}
\label{dgmax_bound_rad}
\nonumber
    \max\limits_{\eta \leq \pi/\alpha}\Bigg(d_g\frac{1+\mt{cos}((\eta-1)\alpha)}{(\pi-(\eta-1)\alpha)}\Bigg) &\leq \min\limits_{\hat{\vb{u}}_r^k\cdot\hat{\vb{l}} \geq \mt{cos}\varphi}\Bigg(d_r\frac{\mt{sin}\varphi}{\varphi}\hat{\vb{u}}_r^k\cdot\hat{\vb{l}}\Bigg)\\
    d_g &\leq \frac{\pi}{2}d_r\frac{\mt{sin}\varphi}{\varphi}\mt{cos}\varphi
\end{align}
Eq.~\eqref{dgmax_bound_tan} and Eq.~\eqref{dgmax_bound_rad} determine an upper bound on the maximum step size of a target as given by condition (1).

{In Fig.~\ref{fig:rho_alpha} we show how the ratio of the step-size between a target and robot, $\lambda$, changes with the number of sensors $p$ and $\angle \alpha$ (which is proportional to how well the robots disperses in the primary orbit). As derived above, $\lambda = \mt{min}\bigg(\frac{\pi}{2},\textrm{  }\frac{\alpha}{\mt{sin}(\pi-\alpha)}\bigg)\frac{\mt{sin}(2\pi/p)}{2\pi/p}$. For a given number of sensors on a robot, $\lambda$ increases with an increase in $\angle\alpha$ until $\pi/2  > \frac{\alpha}{\mt{sin}(\pi-\alpha)}$. We can also see that with an increase in the number of sensors on a robot, $p$, the ratio of the step size of a target to a robot increases and tends to $\pi/2>1$, that is, we can guarantee convergence (encapsulation) even when the target moves faster than the robots in the swarm. Previous approaches in the literature typically assume that the target moves slower than the robots \cite{xue2011swarm,hollinger2009efficient,li2022cooperative,hung2016scalable, DQ_catchMovingTarget}.
% \hkc{add citation}.
%That is, as compared to the literature our control algorithm can guarantee the encapsulation of an escaping target that is moving faster than an individual robot in the swarm.
}
% In Fig. \ref{fig:rho_alpha} we show that for a given total number of sensors $p$ on a robot, with an increase in $\angle \alpha$ (which is proportional to how well the robots disperses in the primary orbit), the ratio of the step-size between a target and robot increases and tends to $\pi/2$
%\hkc{the previous sentence does not describe the figure. as P grows, the ratio goes to pi/2, not with alpha. need more explanations regarding the shape of the curve}. \hkc{also, does this mean the we can have the target move faster then the robot? if so, state that explicitly. it is a cool result, especially compared to the literature}

The increased accuracy in the estimation of the relative location of nearby robots and target enables the robot to disperse quickly in the primary orbit with less chattering. 
Apart from the more accurate estimation, with an increase in total sensors on a robot, the sensing radius $\beta_r$ of a robot increases (Eq.~\eqref{bound_beta_r}), which enables quick dispersion because of a robot's behavior of remaining outside other robots' influence region.
{This results in blocking the escaping paths of the target efficiently.}
The ratio between the target and robot step-sizes is zero when the robot has less than three sensors, for all values of $\angle\alpha$, implying that a minimum of three sensors are required to encapsulate a moving target. 
\begin{figure}
  \centering
  \includegraphics[width=1\linewidth]{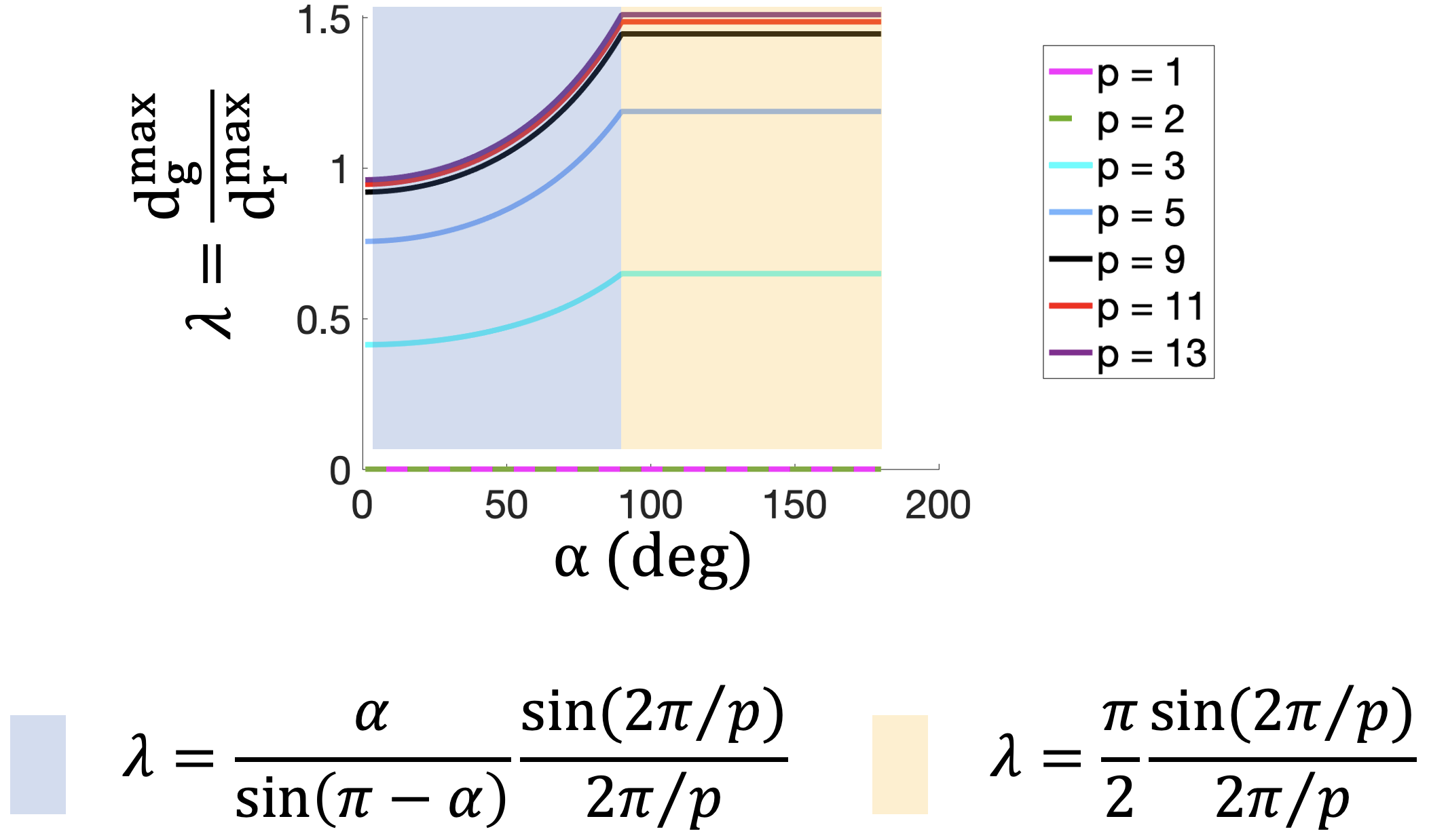}
  \caption{{For a given number of sensors on a robot $p$, $\lambda$ increases with an increase in $\angle\alpha$ until $\pi/2  > \frac{\alpha}{\mt{sin}(\pi-\alpha)}$. }The ratio between the step-size of a target and robot tends to $\pi/2$ with an increasing $p$ {indicating that the swarm can encapsulate a target moving faster than the individual robots in the swarm}. 
%\hkc{this figure is confusing. p=1 an p=11 look like the same color so i did not understand why p=1 was above. why is the graph not smooth? why does it flatten out at 90? also, why is it $\rho$ and not $\lambda$ as in the lemma?}  
}
  \label{fig:rho_alpha}
\end{figure}
\newline

\noindent\textbf{Absence of livelocks and encapsulation of the target $g$:} 
{Similar to Lemma~\ref{lemma_rencap}, we can compute the total number of robots, $n_i$, that can be simultaneously present in an $i^{\mt{th}}$ orbit. If at a time step $T$ there are less than $n_i$ robots in $Or_i$, empty spots that could potentially be occupied by nearby robots, will be present in this orbit.
As discussed in Section~\ref{section_orbits},} the robots in the influence of a target either move toward the target or move, typically, in opposite tangential directions in adjacent orbits. 
This ensures that a dynamic empty spot %\hkc{need to explain what you mean by "dynamic empty spot"} 
present in an orbit $Or_i$ and the robots present in $Or_{i+1}$ move so as to align with each other. 
As we proved above, the robots present in the encapsulation ring (or the primary orbit) are guaranteed to continuously orbit the target.
So, when there are less than $n_g$ robots in the encapsulation ring, a dynamic empty spot is present in the primary orbit which will be eventually occupied by a robot orbiting in the secondary orbit $Or_1$. When either all the empty spots in the primary orbit are filled by the robots or there are at least $n_g$ robots in it, a target will be encapsulated.
Once that happens we have from assumption (\ref{targetAssum}) that the target will stop emitting its signal and set its control parameters to zero thereafter. All the robots that were in secondary orbits and in the influence of this target will transition into random walk behavior. Hence assumption (\ref{targetAssum}) ensures that the robots would not be stuck in the secondary orbits of an encapsulated target and can transition into target-searching behavior after one target is encapsulated. 
% \newline

% \noindent\textbf{Absence of livelocks:} The robots tend to move in opposite tangential directions in adjacent orbits. This ensures that a dynamic empty spot present in an orbit $Or_i$ and the robots present in $Or_{i+1}$ move so as to align with each other. This process of occupying empty spots in consecutive orbits continues until all the spots in the primary orbit are occupied by the robots. 
\end{proof}

\begin{lemma}
\label{lemma_randTar}
{Consider a swarm with a total of $n$ robots and }a target $g \in \mathcal{G}$ moving randomly in the bounded environment (as described by motion model~\ref{randMotion} in Section~\ref{section_targetMotion}). If,
\begin{enumerate}
    \item the inner radius of the primary orbit $Or_0^{\mt{inner}} \geq r_g^{\mt{safe}} + d_r^{\mt{max}}$
    % \item the inner radius of the primary orbit $$Or_0^{\mt{inner}} \geq \sqrt{(r_g^\mt{safe})^2 + r_r^2 - 2r_rr_g^\mt{safe}\mt{cos}(\pi/p)} + r_r + d_g^{\mt{max}}$$
    \item the maximum step size of the target, $d_g^{\mt{max}} \leq \lambda d_r^{\mt{max}}$ where $$\lambda = \Bigg(n-\Bigg\lfloor \frac{2\pi}{\mt{cos}^{-1}\bigg(1 - \frac{(\beta_r + r_r)^2}{2(Or_0^{\mt{inner}})^2}\bigg)} \Bigg\rfloor+1\Bigg)^{-1}$$
\end{enumerate}
the target $g$ will be encapsulated eventually.
% \hkc{what is $n$?} \himani{$n$ is total number of robots in the swarm. It was defined in the problem statement subsection.}
\end{lemma}
\begin{proof}
The challenge in encapsulating a randomly moving target is in ensuring that robots avoid colliding with a target. 
For example, say at time $T$ a robot $i$ is in the primary orbit sandwiched between a target on one side at a distance of $ r_g^{\mt{safe}}+d_g^{\mt{max}}$ and a robot $j$ at a distance of $r_r^{\mt{safe}}$ on the other side, along the target's LOS vector. 
Furthermore, consider that for time steps $T$ until $T + 2$, the randomly moving target acts adversarial by trying to collide with the robot. That is, at every time step it moves towards robot $i$.
% \amy{i wouldn't call this adversarial behavior}.

To ensure that the $i^{\mt{th}}$ robot avoids colliding with the target at $T+1$, it must choose a heading direction $\gamma_i \in \Theta_g^{\mt{avo}}$. 
% \amy{i started making grammar edits, but actually this whole paragraph sounds a bit weird. maybe take a read through and change some of the phrasing} 
However, due to the presence of robot $j$, it cannot move a nonzero distance at time $T$ in the intended heading direction. Hence the robot would violate the target-robot safety specification at $T+1$. 
Now, at $T$, say the $j^{\mt{th}}$ robot had moved away from robot $i$, implying that at $T+1$, the $i^{\mt{th}}$ robot will not sense the $j^{\mt{th}}$ robot and will be free to move away from the target. That is, it took a minimum of two time steps for the $i^{\mt{th}}$ robot to move away from the target. Hence, to ensure safety for this scenario, $d_g^{\mt{max}} \leq d_r^{\mt{max}}/2$ and $Or_0^{\mt{inner}} \geq r_g^{\mt{safe}} + d_r^{\mt{max}}$ (condition 1).

Generalizing this, let the total robots present in the environment be $n$ and $\widetilde{n_0} = \Bigg\lfloor \frac{2\pi}{\mt{cos}^{-1}\bigg(1 - \frac{(\beta_r + r_r)^2}{2(Or_0^{\mt{inner}})^2}\bigg)} \Bigg\rfloor$ 
% \amy{why do you have to define a new variable? can't you just use $\lfloor n_0 \rfloor$ } 
be the number of robots that can be simultaneously present in the primary orbit marginally outside $Or_0^{\mt{inner}}$ without repelling each other. Then, in the worst case scenario, the total time steps for which a robot present on $Or_o^{\mt{inner}}$ may have to remain idle ($d_r=0$) is ($n-\widetilde{n_0}+1$).  
This follows from the fact that only ($n-\widetilde{n}_0$) number of robots contribute to the idle waiting time of a robot in the primary orbit present on $Or_0^{\mt{inner}}$.
This constraint on the idle time of a robot in the primary orbit gives us an upper bound $\lambda$ (condition 2) on how slow an adversarial target needs to be with respect to a robot to ensure safety.
The analysis for stability and encapsulation follows from Lemma~\ref{lemma_randTarEscape}.
\end{proof}

\begin{lemma}
\label{lemma_predeterTarEscape}
Consider a target $g \in \mathcal{G}$ moving in an unknown pattern until it senses a robot in its escape domain (as described by motion model~\ref{predMotion} in Section~\ref{section_targetMotion}). If,
\begin{enumerate}
    \item the maximum step size of the target when moving in an unknown pattern, $d_g^{\mt{max}} < \lambda d_r^{\mt{max}}$ where 
    \begin{align*}
        \lambda &= \frac{\mt{sin}\varphi}{\varphi}\mt{cos}(\varphi), \quad
        \\
        \varphi &= \pi/p, \quad \mt{for\textrm{ }symmetric\textrm{ }sensor\textrm{ }placement}
        % \varphi = \pi/p  \textrm{ for symmetric sensor placement}
    \end{align*}
    % \amy{you may want to use a different variable. I thought this was a v for velocity at first}
    \item the maximum step-size of the target when escaping nearby robots, 
    \begin{align*}
        \lambda &= \mt{min}\bigg(\frac{\pi}{2},\textrm{  }\frac{\alpha}{\mt{sin}(\pi-\alpha)}\bigg)\frac{\mt{sin}\varphi}{\varphi}\mt{cos}\varphi \\
        \alpha &= \mt{cos}^{-1}\bigg(1 - \frac{(\beta_r + r_r)^2}{2(r_g^{\mt{escape}})^2}\bigg) \\
        \varphi &= \frac{\max(\phi^k - \phi^{k+1})}{2},\quad k = \{1 \cdots p\}\\
        &= \pi/p, \quad \mt{for \textrm{ }  symmetric \textrm{ } sensor \textrm{ }  placement}
    \end{align*}
\end{enumerate}
\end{lemma}
the target $g$ will be encapsulated eventually.

\begin{proof}
This scenario is comparable to hunting problems \cite{wu2015hunting,hamed2020improvised} 
% \hkc{add citation} 
where the target moves at slower speeds in some unknown motion pattern. But as soon as it detects (target's sensing limited to $r_g^{\mt{escape}}$) a predator (robot) in its domain, it escapes at a faster speed than the predator. 
To ensure that robots in the secondary orbits move toward the target, we require that,
% $\mathbb{E}(u_{gr}^{rad})\leq 0$. Using Eq. ~\eqref{relVel_gr_rad} and Eq. ~\eqref{avg_ur},
\begin{align}
\nonumber
\label{predeter_rad}
    \mathbb{E}[\vb{u}_g \cdot \hat{\vb{l}}] - \mathbb{E}[\vb{u}_r \cdot \hat{\vb{l}}] < 0\\
    d_g(\hat{\vb{u}}_g \cdot\hat{\vb{l}}) < d_r\frac{\mt{sin}\varphi}{\varphi}\mt{cos}\varphi
\end{align}
Eq.~\eqref{predeter_rad} is always satisfied if $d_g < d_r\frac{\mt{sin}\varphi}{\varphi}\mt{cos}\varphi$. It is trivial to show using Eq.~\eqref{relVel_gr_tan} and Eq.~\eqref{avg_ur} that the constraint $d_g < d_r\frac{\mt{sin}\varphi}{\varphi}\mt{cos}\varphi$ also ensures that robot in a primary orbit will encircle the target. 
As shown in Lemma~\ref{lemma_randTarEscape}, the robots can successfully encapsulate an escaping target as long as a target step-size is within the bounds given by Eq.~\eqref{dgmax_bound_rad} and Eq.~\eqref{dgmax_bound_tan}.
% This constraint also satisfies Eq. ~\eqref{dgmax_bound_rad} and Eq. ~\eqref{dgmax_bound_tan}, that is the robots will successfully encircle an escaping target. 

\end{proof}
% \amy{you should mention this way earlier when you first introduce the algorithm.}

%%%%%%%%%%%%%%%%%%%%%%%%%%%%%%%%%%%%%%%%%%%%%%%%%%%%%%%%%%%%%%%%%%%%%%%%

\section{Simulation Results}
\label{sec_sim}
In this section, we study the effect of the total number of sensors $p$, target-robot step-size ratio $\lambda$, and noisy sensors  on the global behavior of the swarm. 
For each case, we consider three different target motion models: (i) target performs random walk in the environment, (ii) target perform random walk until there exists a robot such that $r_{gr} \leq r_g^{\mt{escape}}$, (iii) target moves with constant velocity until there exists a robot such that $r_{gr} \leq r_g^{\mt{escape}}$.
% \amy{this third scenario doesn't match the third scenario that you've mentioned before. or are you assuming here that constant velocity is the "unknown pattern"?}.
The simulation environment consists of one moving target and ten robots. The total time is capped at 4000 time-steps. 
Due to the inherent randomness in the motion of the robots and targets, we ran 50 simulations for each data point with the same initial conditions. All the robots were initialized arbitrarily such that they lie in a sector of $\pi/4$ with respect to the target's center. 
% a quadrant
% \amy{i don't get what this means}. 
This is done to show the ability of the swarm to successfully encapsulate a target when it has all the escape paths open. 
\newline

\begin{figure}
  \centering
  \includegraphics[width=\linewidth]{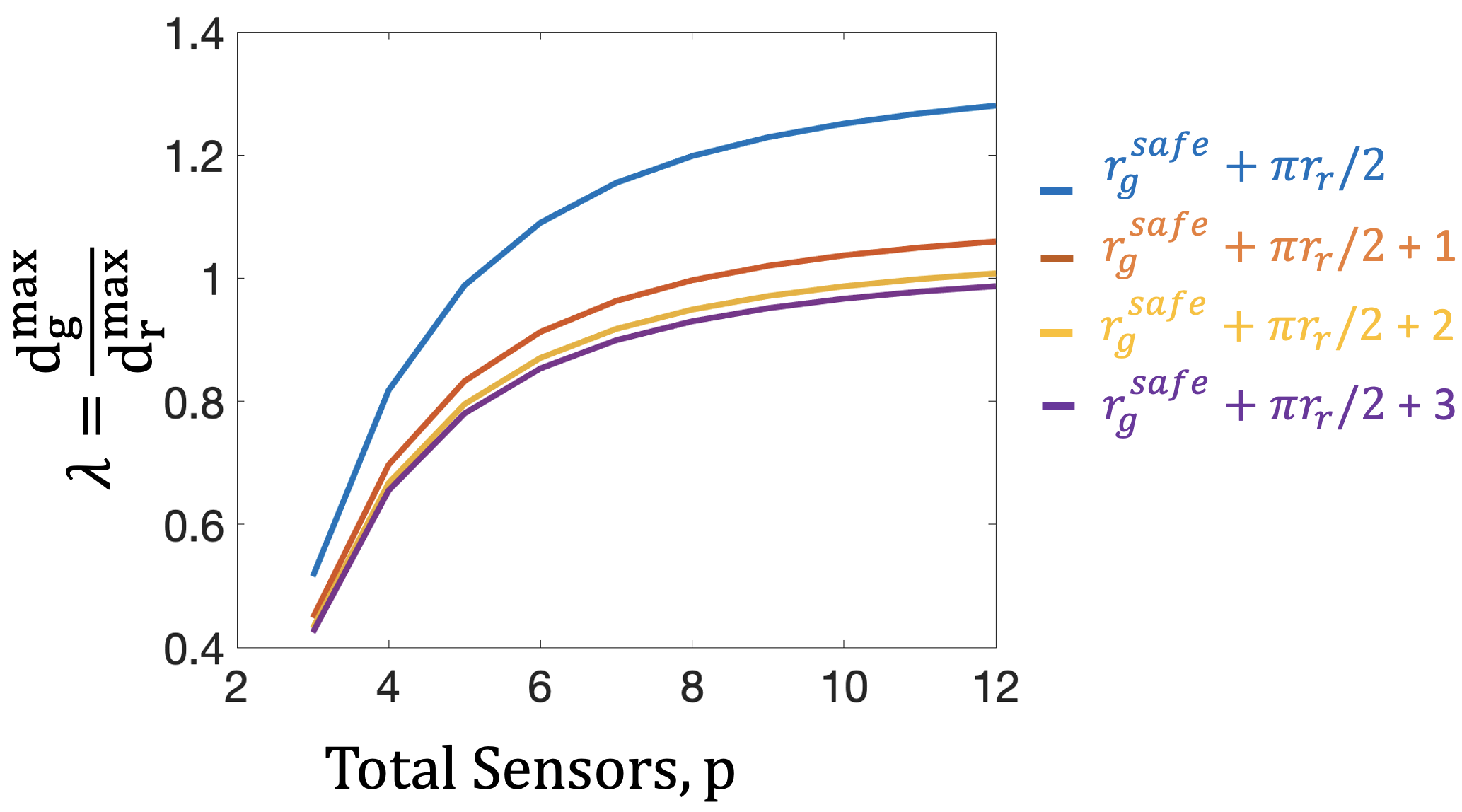}
  % \vspace{-1.3em}
  \caption{For a given $p$, the ratio between the step-size of a target and robot decreases with an increase in $r_g^{\mt{escape}}$, indicating that as the target is able to detect robots sooner, to ensure encapsulation it must also move slower.
%   as a function of the angle subtended at the target's center by adjacent robots present on $Or_0^\mt{inner}$.
% \hkc{$\rho$ or $\lambda$?}  
}
  \label{fig:rescapeg_rho_p}
  % \vspace{-1.3em}
\end{figure}

\begin{figure}[t!]
    \centering
    \begin{subfigure}{0.75\linewidth}
    \includegraphics[width=\textwidth]{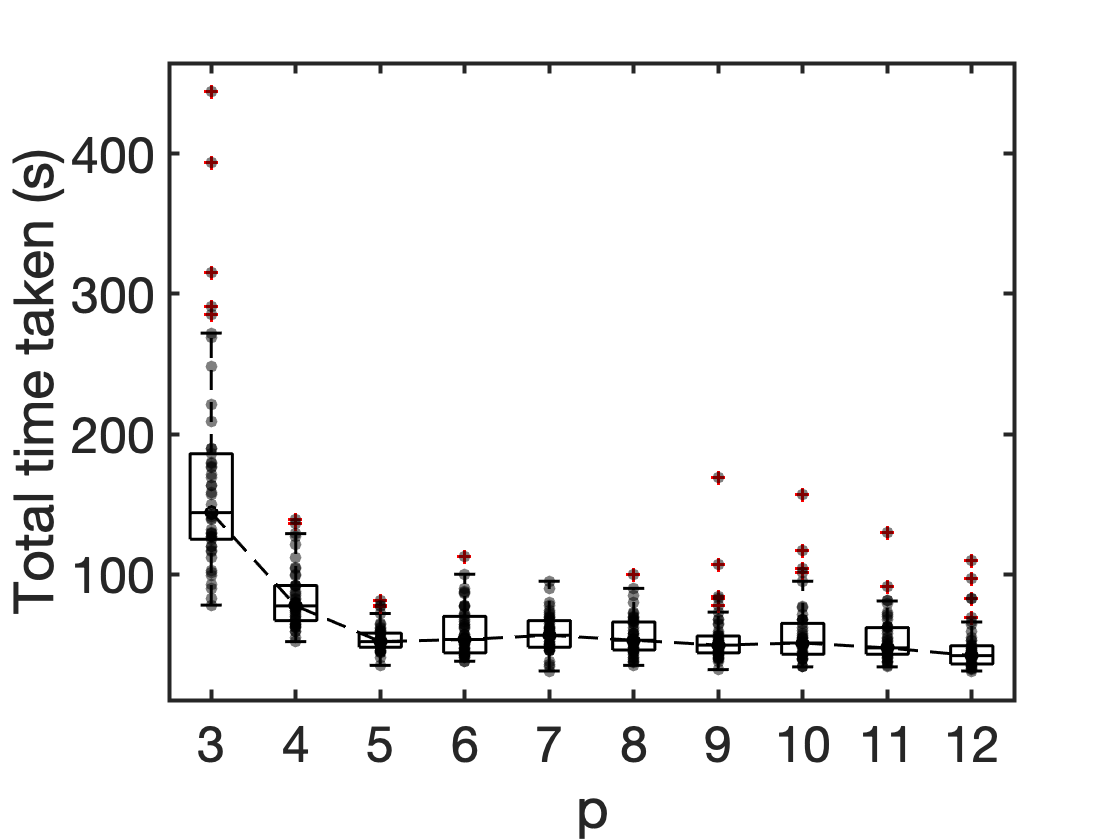}
     \caption{}
      \label{fig:asyncSync_RW}
    \end{subfigure}
    \begin{subfigure}{0.75\linewidth}
    \includegraphics[width=\textwidth]{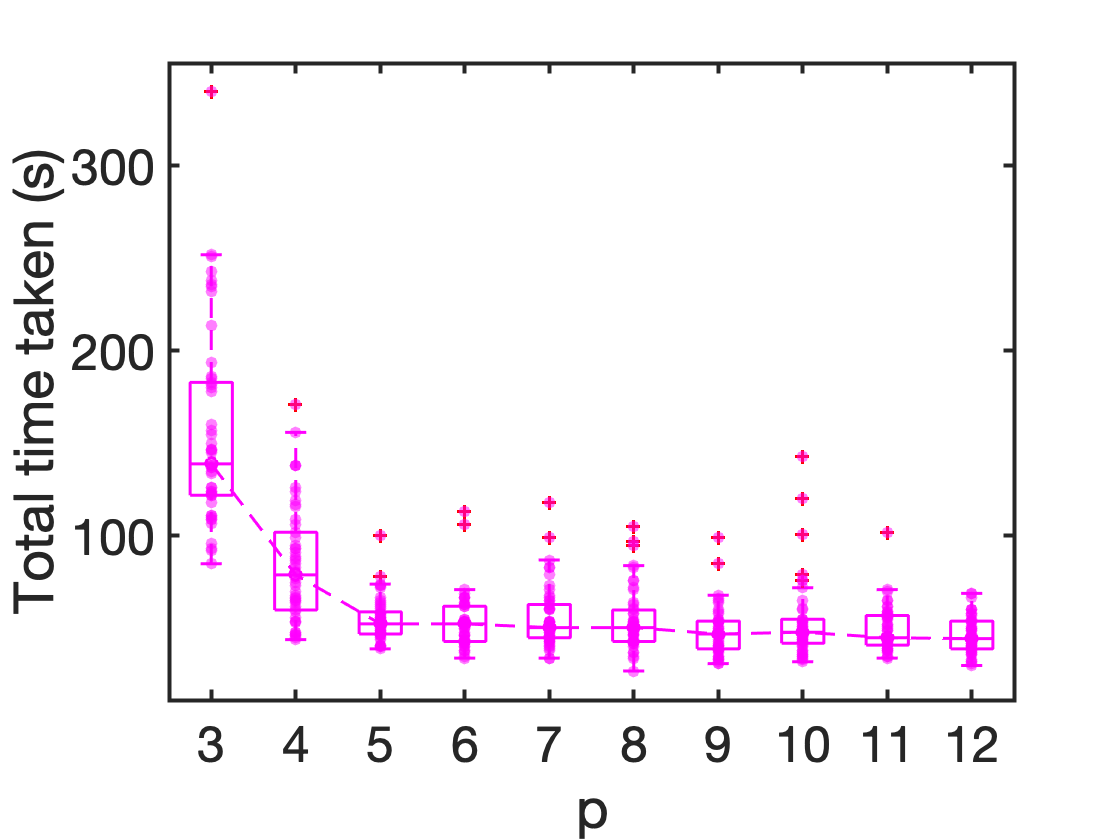}
     \caption{}
      \label{fig:asyncSync_RWesc}
    \end{subfigure}
    \begin{subfigure}{0.75\linewidth}
    \includegraphics[width=\textwidth]{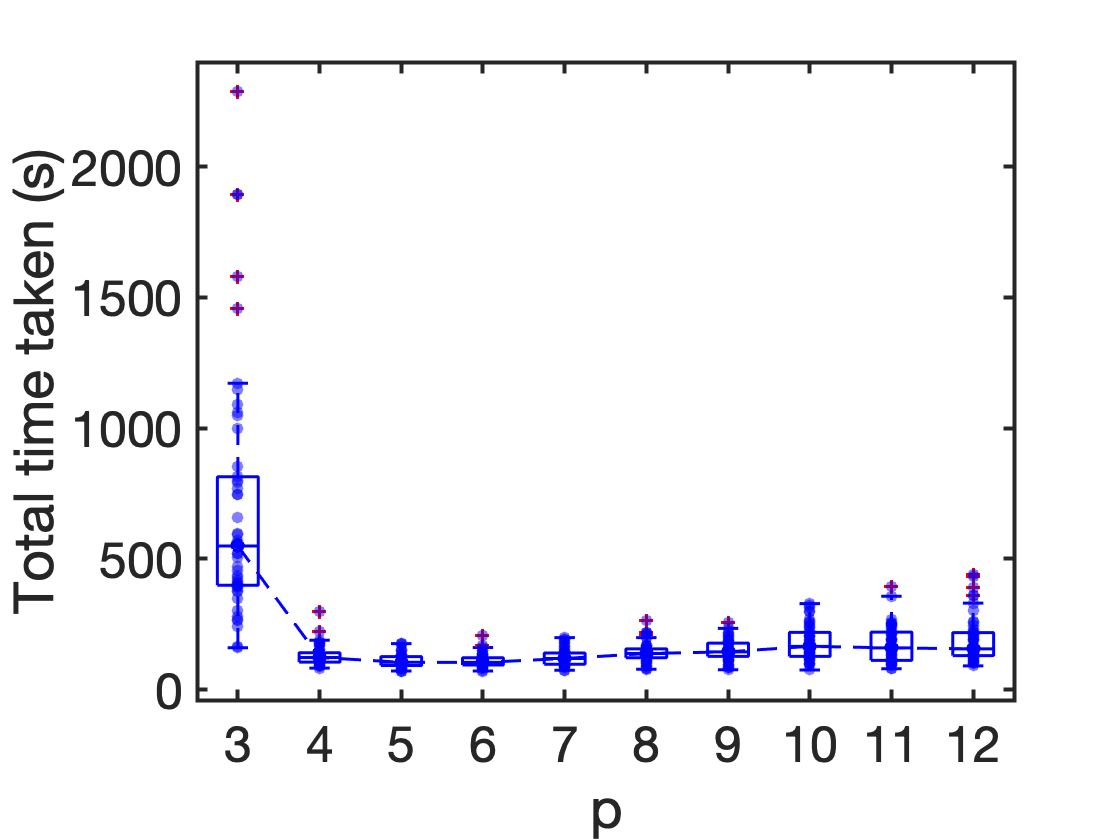}
     \caption{}
      \label{fig:asyncSync_const}
    \end{subfigure}
    % \vspace{-0.4em}
    \caption{The total time taken for task completion as a function of $p$ such that (a) target performs random walk in the environment (b) target performs random walk while escaping from nearby robots (c) target moves with a constant velocity while escaping from nearby robots. 
    % For asynchronous execution, the upper bound on the maximum step size of the target is reduced by the maximum phase difference between robots.
    The box plot shows median, $25^{\mt{th}}$ and $75^{\mt{th}}$ percentiles and the min/max values. The line connects the medians.}
\label{fig:async_sync}
\end{figure}

\noindent\textbf{Effect of the total number of sensors:} 
For a given $p$, with an increase in the radius of the escape domain of a target, the bound on the maximum step size of a target  decreases as shown in Fig.~\ref{fig:rescapeg_rho_p}. This is because, with an increase in $r_g^{\mt{escape}}$, a target gets a higher margin for escaping.
{From Lemmas~\ref{lemma_randTarEscape} - ~\ref{lemma_predeterTarEscape}, the ratio $\lambda$, and hence the target's step size, is dependent on the total number of sensors on a robot, $p$, and $r_g^{\mt{escape}}$. 
When the escape domain of the target, $r_g^{\mt{escape}} \leq r_g^{\mt{safe}} + \frac{\pi}{2}d_r^{\mt{max}}$, an increase in the total number of sensors on a robot enables the swarm to capture a faster moving target. 
This can be seen in Fig.~\ref{fig:rescapeg_rho_p}, where the blue line corresponds to $r_g^{\mt{escape}} = r_g^{\mt{safe}} + \frac{\pi}{2}r_r$. As $p$ increases, $d_r^{\mt{max}}$ tends to $r_r$ and $\lambda$ becomes greater than 1. 
If the escape domain is further increased, that is $r_g^{\mt{escape}} > r_g^{\mt{safe}} + \frac{\pi}{2}d_r^{\mt{max}}$, then $d_g^{\mt{max}}$ decreases proportionally because a robot's step size is limited to $d_r^{\mt{max}}$.  } 
%\hkc{add some comments about when we allow the target to move faster than the robot.} 
% On the other hand, an increase in the total number of sensors on a robot enables the swarm to capture a faster moving target for a given $r_g^{\mt{escape}}$ because an increase in $p$, decreases the sensor resolution error and a robot has a better estimate of target's location. 
For a given $n_g$, $r_g^{\mt{escape}}$ and $r_g^{\mt{encap}}$, Fig.~\ref{fig:async_sync} shows how varying the total number of sensors, and hence the target-robot step-size ratio $\lambda$, affects the total time taken for target encapsulation for each type of target motion model.
\newline

\begin{figure}[t!]
    \centering
    \begin{subfigure}{0.9\linewidth}
    \includegraphics[width=\textwidth]{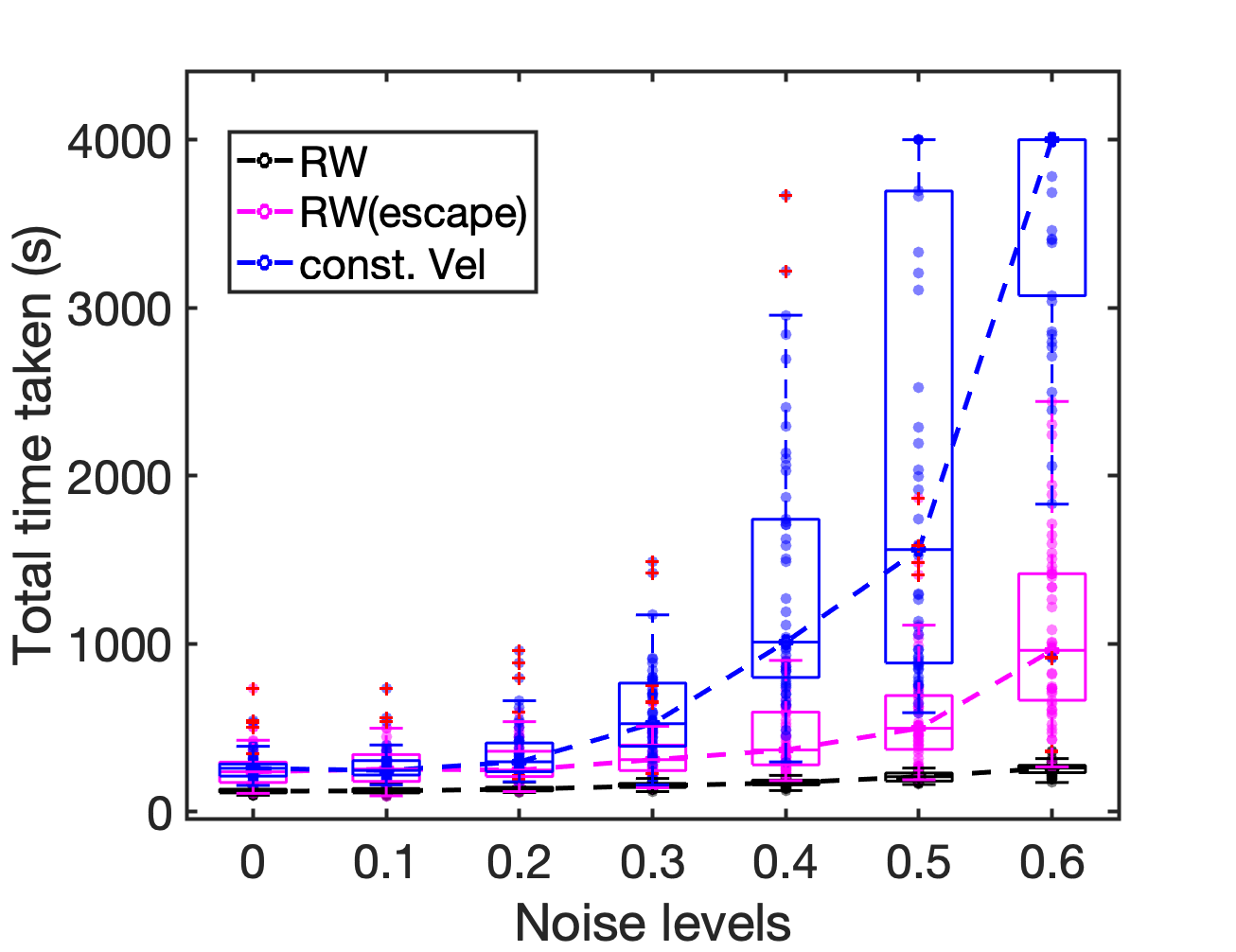}
     \caption{}
      \label{fig:totTime_noise}
    \end{subfigure}
    \begin{subfigure}{0.89\linewidth}
    \includegraphics[width=\textwidth]{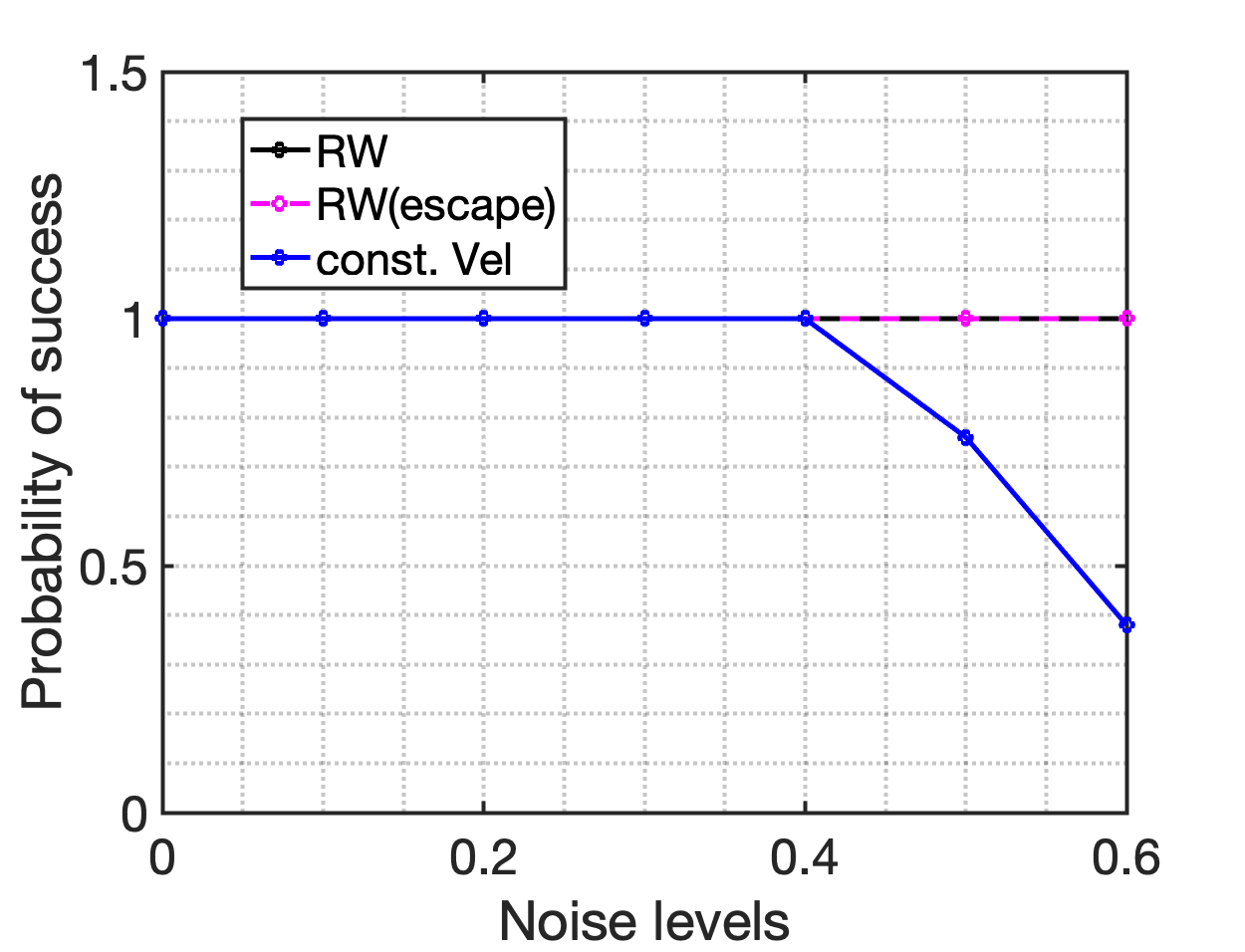}
     \caption{}
      \label{fig:prob_noise}
    \end{subfigure}
    \caption{(a) The total time taken for task completion as a function of noise levels for $p = 7$, $\lambda = 1.1549$ (total time capped at 4000 time-steps) and different target motion models (b) probability of success for task completion
    % \hkc{need to fix box colors}
    }
\end{figure}

\noindent\textbf{Effect of noisy sensors:}
To study the effect of noise we added Gaussian noise to each sensor reading, $z_s^k = (1 - n_s^k)\sum_{j\in N_s^k} B_s(d^{k}_j),$ $\, n_s^k\sim \mathcal{N}(0,\sigma^2)$ and $n^k_s \leq 1$. 
Similar to the results obtained in our previous work \cite{selfPaper}, for all noise levels, we did not observe any collision within the swarm.
However, to ensure that a robot does not collide with a moving target or the environment boundary, we increase the radius of $Or_0^{\mt{inner}}$ in proportion to the standard deviation of the noise.
Fig.~\ref{fig:totTime_noise} shows the total time taken by the swarm to encapsulate a target with $p = 7$ and $\lambda = 1.1549$.  
With an increase in noise level, a robot's estimate of the target's location becomes less accurate, leading to an increase in the total time taken for encapsulation. Furthermore, as can be seen in Fig.~\ref{fig:prob_noise} for noise levels greater than 50\%, the probability of success for target encapsulation drops to 40\% when a target moves with constant velocity. 
\newline

\begin{figure}[!ht]
  \centering
  \includegraphics[width=0.9\linewidth]{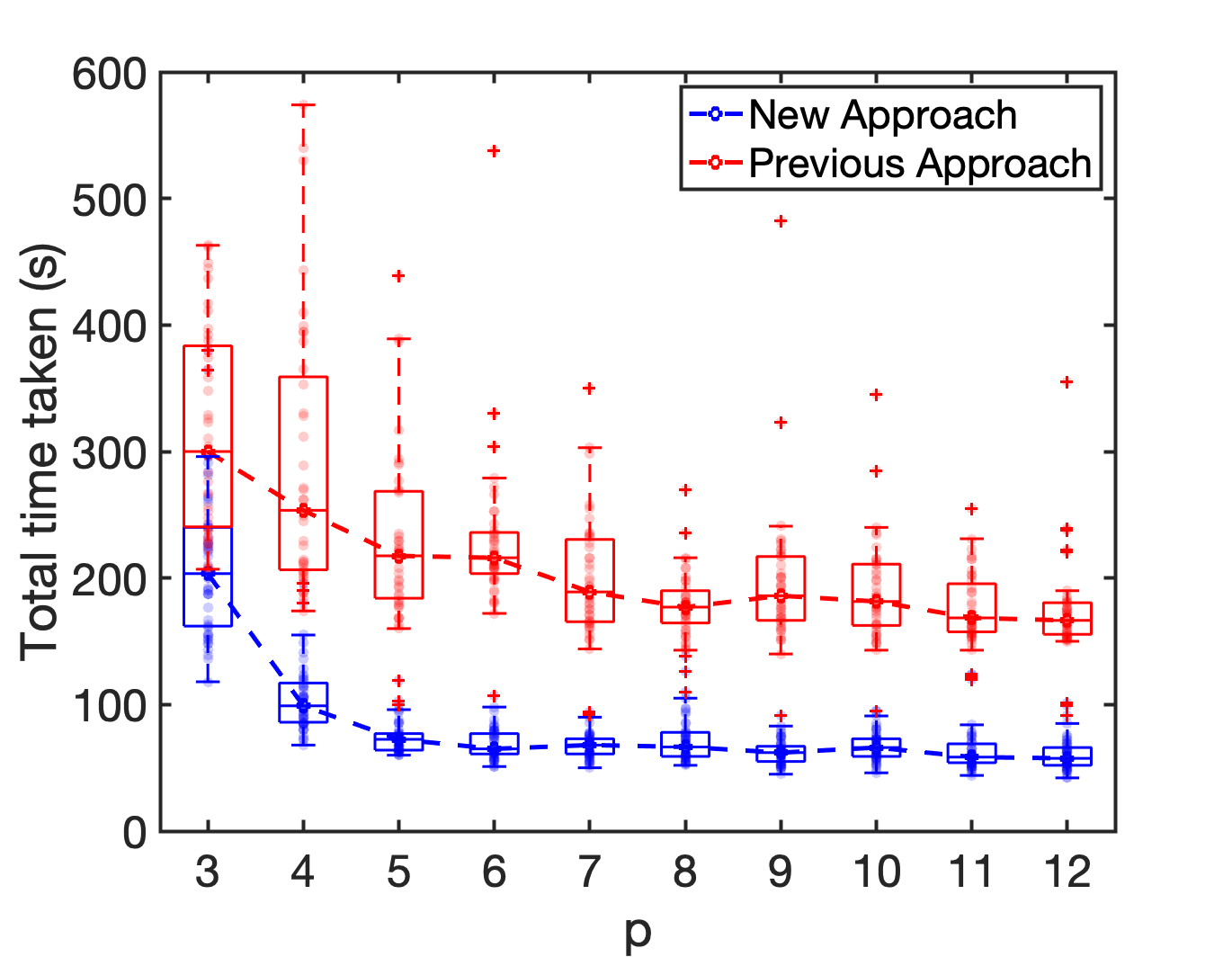}
  % \vspace{-1.3em}
  \caption{For a given $p$, the total time taken to encapsulate a static target is lower for the new approach introduced in this paper as compared to our previous approach in \cite{selfPaper}.
  }
  \label{fig:comparison}
\end{figure}

\noindent\textbf{Comparison with algorithm in~\cite{selfPaper}:}
The algorithm we proposed in this paper is more efficient in terms of the total time taken by the swarm to encapsulate a static target as compared with our previous method in \cite{selfPaper}. This is due to the orbiting behavior of the swarm when a robot cannot move toward the target which results in a faster occupancy of empty spots in the encapsulation ring. This is shown in Fig.~\ref{fig:comparison}.
\newline

\noindent\textbf{Scalability:} In the supplementary video, we run additional simulations to show the effect of asymmetric sensor placement,
the validity of our algorithm for non-circular robots and demonstrate the scalability of our algorithm with a large-scale simulation of 120 robots and 15 targets moving with different motion models. 
% \amy{you say you investigated - what were the findings?}

%%%%%%%%%%%%%%%%%%%%%%%%%%%%%%%%%%%%%%%%%%%%%%%%%%%%%%%%%%%%%%%%%%%%%%%%
\section{Conclusion}
In this paper, we propose a decentralized scalable algorithm for a minimalist swarm to encapsulate dynamic targets with unknown motion without requiring the exact knowledge of the relative positions or memory of the previous control inputs. We consider different scenarios of target motion and compute bounds on the target-robot step-size ratio to provide convergence guarantees. We observed the emergence of robots maintaining an approximate phase difference of $2\pi/n_g$ in the encapsulating ring, resulting in uniform distribution around the target and hence closing off its escaping directions. Furthermore, using extensive simulations we studied the effect of noisy sensors and showed the validity of our algorithm for non-circular robots. 
Our controller can be generalized for robots equipped with non-isotropic sensors which are not accurate in measuring the relative distances between two entities. If the bounds on the measurement error are known, our analysis can be used to compute bounds on the target-robot step-size ratio to ensure guaranteed target encapsulations. 
In the future, we are planning on implementing our algorithm on physical robots. We will also study the trade-off between incorporating the memory of previous states on the desired emergent behavior and providing timing bounds on task completion.  
% - we observe the emergence of maintaining an approximate phase difference of $2\pi/n_g$ in $Or_0$\\
% - We can generalize this control law for the non-isotropic sensors which are not accurate in measuring the relative distances between two entities. 

%%%%%%%%%%%%%%%%%%%%%%%%%%%%%%%%%%%%%%%%%%%%%%%%%%%%%%%%%%%%%%%%%%%%%%%%

%----- references---
\bibliographystyle{IEEEtran}
\bibliography{ref}
\end{document}